\documentclass[letterpaper]{article} 
\usepackage{aaai2026}  
\usepackage{times}  
\usepackage{helvet}  
\usepackage{courier}  
\usepackage[hyphens]{url}  
\usepackage{graphicx} 
\usepackage{natbib}  
\usepackage{caption} 
\usepackage{algorithm}
\usepackage{algorithmic}

\usepackage{array}      
\usepackage{caption}
\usepackage{multirow}
\usepackage{booktabs}
\usepackage{colortbl}  
\usepackage{hhline} 
\usepackage{amsmath} 
\usepackage{makecell}
\usepackage{xurl}
\usepackage{graphicx}
\usepackage{subcaption}

\usepackage[capitalise]{cleveref}

\newlength\savewidth\newcommand\shline{\noalign{\global\savewidth\arrayrulewidth
  \global\arrayrulewidth 1pt}\hline\noalign{\global\arrayrulewidth\savewidth}}
\newcolumntype{L}[1]{>{\raggedright\arraybackslash}p{#1}}  

\newcommand{\ourmethod}{\textit{LLM2CLIP}}

\usepackage[table]{xcolor}
\selectcolormodel{cmyk}
\definecolor{myblue}{cmyk}{0.15,0.05,0,0}

\crefname{figure}{Figure}{Figures}
\Crefname{section}{Section}{Sections}
\Crefname{table}{Table}{Tables}

\usepackage[symbol]{footmisc}
\usepackage{lineno}

\usepackage{newfloat}
\usepackage{listings}
\DeclareCaptionStyle{ruled}{labelfont=normalfont,labelsep=colon,strut=off} 
\floatstyle{ruled}
\newfloat{listing}{tb}{lst}{}
\floatname{listing}{Listing}
\title{LLM2CLIP: Powerful Language Model Unlocks Richer Cross-Modality Representation}

\author{
    Weiquan Huang\textsuperscript{\rm 1}\equalcontrib,
    Aoqi Wu\textsuperscript{\rm 1}\equalcontrib,
    Yifan Yang\textsuperscript{\rm 2}\footnote{Corresponding authors.}\footnote{Project Lead.},
    Xufang Luo\textsuperscript{\rm 2},
    Yuqing Yang\textsuperscript{\rm 2},
    Usman Naseem\textsuperscript{\rm 3},
    Chunyu Wang\textsuperscript{\rm 2},
    Qi Dai\textsuperscript{\rm 2},
    Xiyang Dai\textsuperscript{\rm 2},
    Dongdong Chen\textsuperscript{\rm 2},
    Chong Luo\textsuperscript{\rm 2},
    Lili Qiu\textsuperscript{\rm 2},
    Liang Hu\textsuperscript{\rm 1}\footnotemark[2]
}
\affiliations{
    \textsuperscript{\rm 1} School of Computer Science dnd Technology, Tongji Univeristy, Shanghai, China\\
    \textsuperscript{\rm 2} Microsoft Corporation\\ 
    \textsuperscript{\rm 3} School of Computing, Macquarie University, Sydney, Australia\\
    yifanyang@microsoft.com, lianghu@tongji.edu.cn

}

\usepackage{bibentry}

\begin{document}

\maketitle

\begin{abstract}

CLIP is a seminal multimodal model that maps images and text into a shared representation space by contrastive learning on billions of image–caption pairs. Inspired by the rapid progress of large language models (LLMs), we investigate how the superior linguistic understanding and broad world knowledge of LLMs can further strengthen CLIP—particularly in handling long, complex captions. We introduce an efficient fine-tuning framework that embeds an LLM into a pretrained CLIP while incurring almost the same training cost as regular CLIP fine-tuning. Our method first “embedding-izes” the LLM for the CLIP setting, then couples it to the pretrained CLIP vision encoder through a lightweight adaptor trained on only a few million image–caption pairs. With this strategy we achieve large performance gains—without large-scale retraining—over state-of-the-art CLIP variants such as EVA02 and SigLIP-2. The LLM-enhanced CLIP delivers consistent improvements across a wide spectrum of downstream tasks, including linear-probe classification, zero-shot image–text retrieval with both short and long captions (in English and other languages), zero-shot/supervised image segmentation, object detection, and used as tokenizer for multimodal large-model benchmarks. \textbf{Code \& Models:} \url{https://aka.ms/llm2clip}.

\end{abstract}



\section{Introduction}




\begin{figure}[ht]
    \centering
    \includegraphics[width=1.0\linewidth]{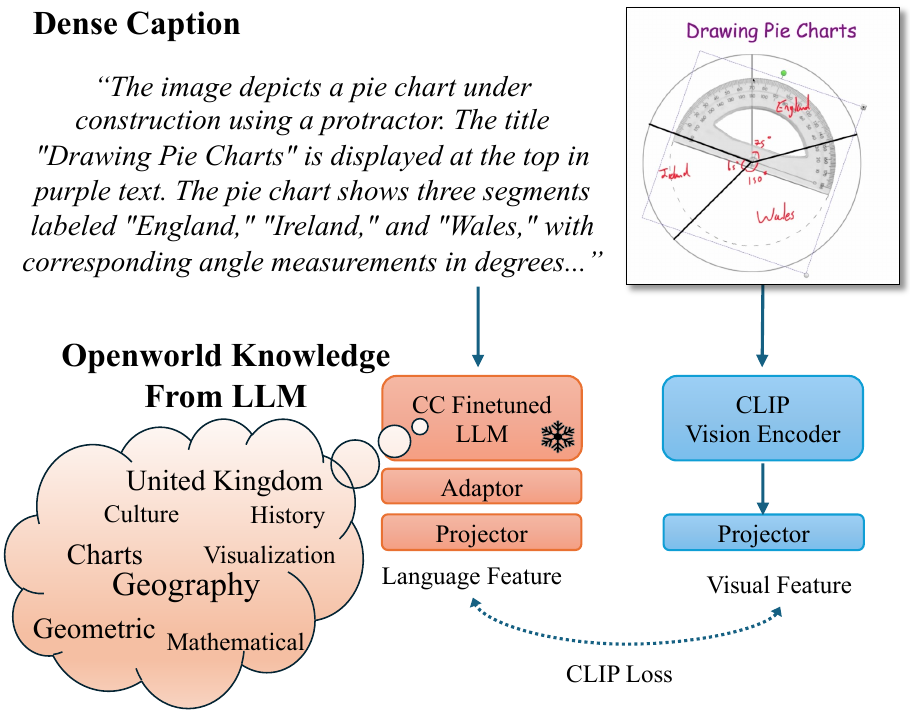}
    \caption{\ourmethod{} Overview. After applying caption contrastive fine-tuning to the LLM, the increased textual discriminability enables more effective CLIP training. We leverage the open-world knowledge and general capabilities of the LLM to better process dense captions, addressing the previous limitations of the pretrained CLIP visual encoder and providing richer textual supervision. 
}
    \label{fig:arch}
\end{figure}


\textit{CLIP}~\cite{radford2021learning,tschannen2025siglip} has emerged as one of the most influential cross-modal foundation models in recent years. 
Trained on hundreds of millions to tens of billions of image--text pairs via contrastive pre-training, it embeds vision and language into a shared representation space. 
As a \emph{retriever}, CLIP underpins zero-shot classification, detection, segmentation, and—most notably—image–text retrieval. 
As a \emph{feature extractor}, it supports a broad spectrum of cross-modal applications, from image/video understanding to text-to-image and text-to-video generation. 
For instance, Multi-modality large language models such as LLaVA~\cite{li2024llavaov} and Qwen-VL~\cite{bai2023qwen} rely on CLIP’s visual features, while image/video generative model like Stable-Diffusion-3~\cite{SD3} and Wan~\cite{kong2024hunyuanvideo} leverage its text encoder. 
Nevertheless, as we push toward broader generality and higher task complexity, the representational capacity inherited from CLIP’s original paradigm is gradually becoming inadequate.

The rapid progress of large language models (LLMs) has dramatically advanced textual understanding and generation, motivating us to inject stronger LLMs into the CLIP training pipeline so as to bolster multimodal capability.  
Prior work such as LLM2Vec~\cite{llm2vec} and NV-Embed-v2~\cite{lee2024nv} has shown that LLMs can be transformed into competitive text–embedding models, now topping leaderboards like MTEB~\cite{muennighoff2022mteb}.  
Follow-up efforts (e.g.\ MM-E5~\cite{chen2025mme5}, VLM2Vec~\cite{Jiang2024VLM2VecTV}) have extended this idea to multi-modality representations.  
Yet CLIP’s light dual-tower architecture remains the de-facto embedding model in many scenarios.  
For instance, in standard zero-shot image–text retrieval, a 400 M-parameter SigLIP-2 So/14 reaches 85.7 (T$\!\to$I) and 94.9 (I$\!\to$T) on Flickr30K at $384{\times}384$ resolution, while VLM2Vec—built on a 7B-parameter LLaVA-1.6 and trained at $1344{\times}1344$—achieves 79.8 and 91.6, respectively.  
CLIP’s lightweight design also scales more readily to large datasets and batch sizes (SigLIP-2~\cite{tschannen2025siglip} is trained on $\sim$40 B image–text pairs). Consequently, mainstream multimodal retrieval systems—and the multimodal encoders that serve VL-LMs and diffusion models—are increasingly built on CLIP-based architectures. This raises a key question: how can we efficiently harness the strengths of modern LLMs to augment these pretrained CLIP models and further elevate their capabilities?

The potential benefits of incorporating LLMs into CLIP are clear. LLMs' strong textual understanding can fundamentally improve CLIP's ability to handle image captions, drastically enhancing its ability to process long and complex texts—a well-known limitation of vanilla CLIP. Moreover, LLMs are trained on a vast corpus of text, possessing open-world knowledge. This allows them to expand on caption information during training, increasing the efficiency of the learning process. 

Bringing LLMs into the CLIP framework, however, raises two key challenges: \textbf{1. Feature separability.}  Vanilla LLM embeddings are not sufficiently discriminative for contrastive training; existing “LLM-as-embedding” recipes target pure-text tasks.  We therefore devise a CLIP-specific \emph{embeddingization} strategy for LLMs that yields far more separable caption features. \textbf{2. Training cost.}  CLIP pre-training is already expensive; naively fine-tuning an LLM jointly would be prohibitive.  We propose a lightweight \emph{fine-tuning} procedure that injects LLM power into a pretrained CLIP at almost no extra cost.

\begin{figure}[h]
    \centering
    \includegraphics[width=1.0\linewidth]{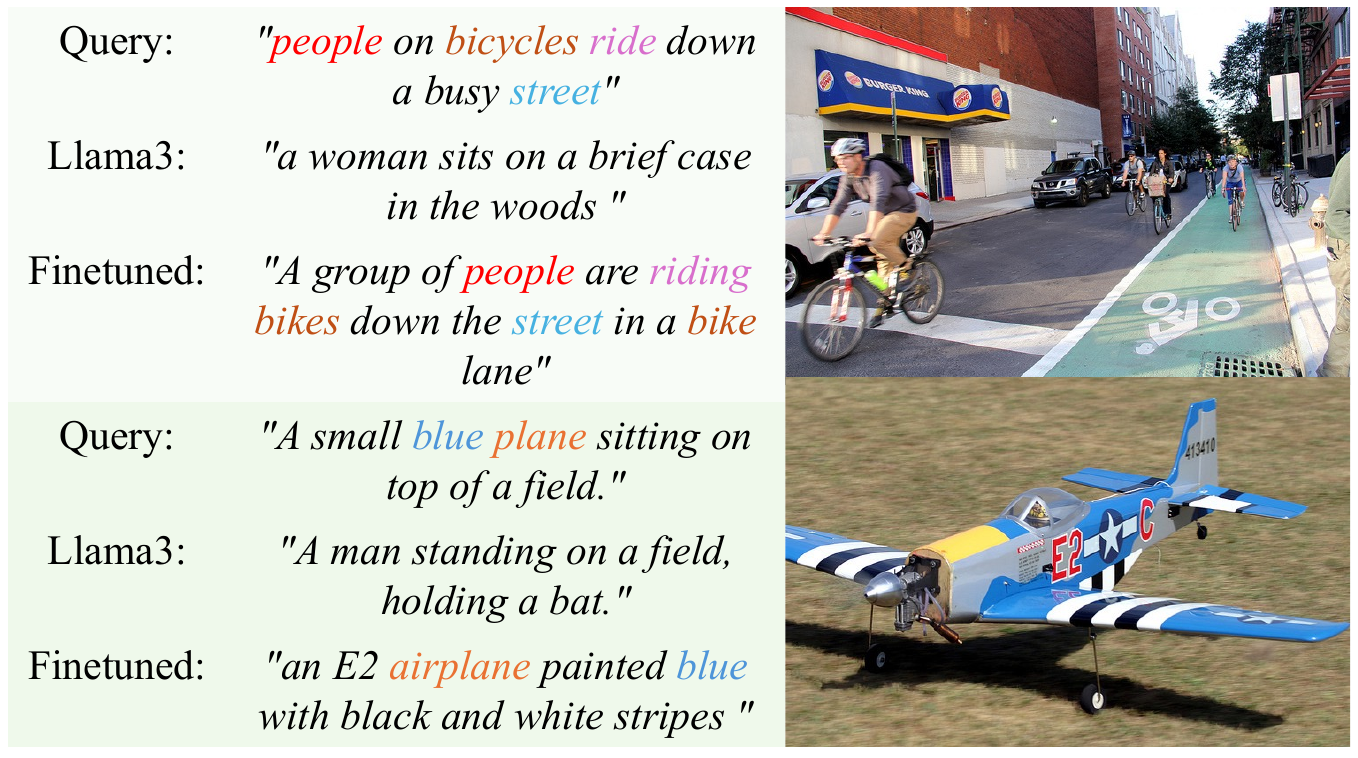}
    \caption{Real examples of top-1 results from the caption-to-caption retrieval experiment in MS COCO 5K test set. Before fine-tuning, Llama3’s results were often unrelated.}
    \label{fig:llm2vec}
\end{figure}

This paper introduces \ourmethod{}, an efficient fine–tuning framework that augments the feature space of a \emph{pre-trained} CLIP with an \emph{embedding-tuned} LLM, thereby importing LLM capability at \textbf{very low cost}.  
Injecting a \emph{vanilla} LLM into CLIP is problematic: as shown by the cases in \Cref{fig:llm2vec}, a COCO test reveals that raw LLM embeddings have poor separability for image captions.  
To remedy this, we perform \emph{caption-contrastive(CC) fine-tuning} on the LLM with a set of high-quality caption datasets, revisiting several design choices—e.g.\ employing \textbf{average pooling} to aggregate token features, enabling \textbf{bidirectional attention}, and training with a \textbf{supervised SimCSE} contrastive loss.  

Another key challenge when introducing LLMs into CLIP is the presence of two text encoders—the original CLIP text encoder and the newly added LLM—which were never aligned during pre-training. We conduct experiments to explore efficient integration of the LLM into the CLIP architecture. As shown in~\Cref{fig:arch}, we adopt a cost-effective solution: \emph{freeze all LLM gradients}, treat its sentence embeddings as fixed features, and append a small learnable \emph{adaptor} trained with the CLIP \emph{visual encoder}. The original CLIP text encoder is discarded during both training and inference. Ablation studies indicate that alternative strategies—combining CLIP and LLM embeddings via triplet contrastive losses or concatenating features—provide marginal or negative gains compared to simply replacing the CLIP text encoder. This minimal design yields an LLM-enhanced cross-modal space while keeping compute cost nearly identical to standard CLIP fine-tuning.

Empirical results confirm that \ourmethod{} yields substantial improvements—even with only a few million training examples—boosting the performance of the original CLIP across a variety of downstream tasks.
Our contributions are as follows:
\begin{itemize}
    \item We empirically demonstrate that injecting LLM capability into CLIP brings significant performance gains.
    \item We design (i) a caption-contrastive fine-tuning recipe that turns an LLM into a effective embedding model for CLIP, and (ii) a efficient fine-tuning method that couples this embedding with a pretrained CLIP.
    \item Extensive experiments reveal that the resulting \textbf{LLM-enhanced CLIP} markedly improves several state-of-the-art models—including EVA02 and SigLIP-2—on a wide spectrum of multimodal benchmarks: short/long-text and cross-lingual image retrieval, zero-shot classification, detection, segmentation, and even as the visual encoder inside LLaVA-1.5.  Remarkably, \ourmethod{} lifts SigLIP-2’s short-caption retrieval by +1.0/+1.9, long-caption retrieval by +14.8/+15.8, and multilingual tasks by +11.9/+15.2.
\end{itemize}

\section{Related Works}

\noindent\textbf{CLIP meets Stronger Language Models.}
Several works have explored the integration of LLMs into CLIP.
JinaCLIP~\cite{koukounas2024jina} employed Jina-embeddings-v2~\cite{gunther2023jinaembedding2} as the text encoder, which is a BERT variant with 137M parameters, supporting longer texts. Though achieving similar visual performance to EVA-CLIP~\cite{EVA}, its text encoder is far behind ours, limiting the potential benefits from LLMs.
MATE~\cite{jang2024mate} designed a learnable adaptor to bridge the gap between CLIP's text encoder and LLMs. They trained the CLIP visual encoder using LoRA on a small dataset focused on long-text image retrieval tasks. However, they did not recognize the critical issue we propose: the poor separability of LLM feature space, which is insufficient for direct support of CLIP training. This paper presents the first comprehensive study of an efficient strategy for incorporating LLMs into CLIP fine-tuning, thereby boosting performance.

\textbf{CLIP meets Longer Captions.}
CLIP's text embedding is widely recognized as coarse and limited to 77 tokens. Many works have attempted to extend caption length and retrain CLIP accordingly, including DCI~\cite{DCI} with human annotation, LaCLIP~\cite{LaCLIP} using ChatGPT and Bard, DreamLIP~\cite{dreamlip} leveraging ShareCaptioner~\cite{chen2023sharegpt4v}, and Recap-DataComp-1B~\cite{Recaption} using Llama3-trained LLAVA1.5. To handle longer captions, these methods employ workarounds such as summarization~\cite{DCI}, segmentation~\cite{LaCLIP,dreamlip}, or positional encoding finetuning~\cite{zhang2024long}. In comparison, LLM2CLIP leverages LLMs as the text encoder, enabling comprehensive understanding of long and dense captions while utilizing LLMs' open-world knowledge.

\begin{table*}[ht]
\centering
\setlength{\tabcolsep}{5pt}  
\footnotesize 
\begin{tabular}{@{}l c | cc | cc | cc | cc | cc | cc@{}}
\hline
\multirow{2}{*}{\textbf{Method}} & \multirow{2}{*}{\textbf{Res}} & \multicolumn{2}{c|}{\textbf{Flickr}} & \multicolumn{2}{c|}{\textbf{COCO}} & \multicolumn{2}{c|}{\textbf{SG4V}} & \multicolumn{2}{c|}{\textbf{Urban}} & \multicolumn{2}{c|}{\textbf{DOCCI}} & \multicolumn{2}{c}{\textbf{Avg}} \\
 &  & \textbf{I2T} & \textbf{T2I} & \textbf{I2T} & \textbf{T2I} & \textbf{I2T} & \textbf{T2I} & \textbf{I2T} & \textbf{T2I} & \textbf{I2T} & \textbf{T2I} & \textbf{I2T} & \textbf{T2I} \\
\hline
\multicolumn{14}{l}{\textbf{ViT-B/16 (86M)}} \\
\hline
ALIGN~\cite{jia2021scaling} & 224 & 80.6 & 62.2 & 52.0 & 43.2 & 75.9 & 80.6 & 62.2 & 59.1 & 59.7 & 62.1 & 66.1 & 61.4 \\
BLIP~\cite{li2022blip} & 224 & 80.6 & 74.1 & 61.7 & 48.5 & 65.8 & 74.3 & 45.5 & 48.5 & 50.5 & 53.5 & 60.8 & 59.8 \\
Long-CLIP~\cite{zhang2024long} & 224 & 85.8 & 70.6 & 56.9 & 40.9 & 94.8 & 93.5 & 79.1 & 79.1 & 63.1 & 71.4 & 75.9 & 71.1 \\
jina-clip-v2~\cite{koukounas2024jinav2} & 224 & 84.4 & 69.8 & 57.1 & 41.9 & 94.5 & 91.1 & 80.4 & 78.0 & 77.6 & 78.2 & 78.8 & 71.8 \\
SigLIP~\cite{zhai2023sigmoid} & 224 & 89.1 & 74.7 & 65.8 & 47.8 & 85.8 & 83.5 & 62.8 & 62.4 & 70.1 & 70.6 & 74.7 & 67.8 \\
MetaCLIP~\cite{xu2023demystifying} & 224 & 85.6 & 70.8 & 59.3 & 41.3 & 90.4 & 86.6 & 68.9 & 63.3 & 70.9 & 71.5 & 75.0 & 66.7 \\
CLIP~\cite{radford2021learning} & 224 & 82.3 & 62.2 & 52.4 & 33.1 & 84.5 & 79.8 & 67.5 & 53.1 & 60.7 & 57.1 & 69.5 & 57.1 \\
\rowcolor[cmyk]{0.05,0.02,0,0}\hspace{2mm}+\ourmethod{}-15M & 224 & \textbf{91.8} & 80.7 & 64.8 & 52.4 & 97.8 & 97.6 & \textbf{90.1} & \textbf{92.1} & 86.1 & 86.9 & \textbf{86.1} & 81.9 \\
EVA02~\cite{sun2023eva} & 224 & 86.2 & 71.5 & 58.7 & 42.1 & 90.5 & 85.5 & 67.0 & 60.8 & 67.7 & 68.0 & 74.0 & 65.6 \\
\rowcolor[cmyk]{0.05,0.02,0,0}\hspace{2mm}+\ourmethod{}-15M & 224 & 91.2 & \textbf{81.4} & \textbf{66.0} & \textbf{53.9} & \textbf{98.5} & \textbf{98.5} & 88.2 & 91.3 & \textbf{86.8} & \textbf{88.5} & \textbf{86.1} & \textbf{82.7} \\
\hline
\multicolumn{14}{l}{\textbf{ViT-L/14 (307M)}} \\
\hline
Long-CLIP~\cite{zhang2024long} & 224 & 90.0 & 76.2 & 62.8 & 46.3 & 97.2 & 97.3 & 82.5 & 86.1 & 66.5 & 78.6 & 79.8 & 76.9 \\
MetaCLIP~\cite{xu2023demystifying} & 224 & 90.1 & 76.4 & 64.5 & 47.1 & 86.4 & 79.5 & 73.4 & 70.0 & 76.5 & 76.7 & 78.2 & 69.9 \\
L-NV2~\cite{zhang2024assessing} & 224 & 87.9 & 75.8 & 62.4 & 48.7 & 94.8 & 94.9 & 81.5 & 80.2 & 76.5 & 78.9 & 80.6 & 75.7 \\
CLIP~\cite{radford2021learning} & 224 & 85.2 & 65.0 & 56.3 & 36.5 & 84.2 & 83.6 & 68.3 & 55.6 & 63.1 & 65.8 & 71.4 & 61.3 \\
\rowcolor[cmyk]{0.05,0.02,0,0}\hspace{2mm}+\ourmethod{}-15M & 224 & 93.0 & 83.5 & 67.4 & 56.4 & 98.7 & 98.6 & 91.5 & 94.1 & 88.3 & 90.4 & 87.8 & 84.6 \\
EVA02~\cite{sun2023eva} & 224 & 89.7 & 77.3 & 63.7 & 47.5 & 91.9 & 89.3 & 73.3 & 68.5 & 73.5 & 75.0 & 78.4 & 71.5 \\
\rowcolor[cmyk]{0.05,0.02,0,0}\hspace{2mm}+\ourmethod{}-3M & 224 & 94.3 & 84.3 & 65.2 & 56.0 & 98.0 & 98.0 & 91.0 & 94.7 & 87.8 & 90.7 & 87.3 & 84.7 \\
\rowcolor[cmyk]{0.05,0.02,0,0}\hspace{2mm}+\ourmethod{}-15M & 224 & 94.6 & 85.0 & 69.5 & 58.3 & 98.9 & 99.1 & 93.6 & 95.7 & 89.8 & 91.2 & 89.3 & 85.8 \\
\rowcolor[cmyk]{0.05,0.02,0,0}\hspace{2mm}+\ourmethod{}-60M & 224 & \textbf{95.9} & \textbf{85.1} & \textbf{71.7} & \textbf{58.5} & \textbf{99.2} & \textbf{99.2} & \textbf{95.2} & \textbf{97.0} & \textbf{90.1} & \textbf{92.0} & \textbf{90.4} & \textbf{86.3} \\
\hline
CLIP~\cite{radford2021learning} & 336 & 87.7 & 67.0 & 58.0 & 37.1 & 86.2 & 84.0 & 72.8 & 57.0 & 67.4 & 65.7 & 74.4 & 62.2 \\
\rowcolor[cmyk]{0.05,0.02,0,0}\hspace{2mm}+\ourmethod{}-15M & 336 & 91.2 & 82.1 & 65.5 & 53.6 & 98.1 & 98.4 & 90.3 & 93.2 & 87.7 & 89.0 & 86.6 & 83.3 \\
\rowcolor[cmyk]{0.05,0.02,0,0}\hspace{2mm}+\ourmethod{}-60M & 336 & 93.9 & 84.3 & 68.5 & 54.8 & \textbf{98.9} & 99.1 & \textbf{96.6} & \textbf{97.9} & 89.6 & 91.6 & \textbf{91.1} & 84.5 \\
EVA02~\cite{sun2023eva} & 336 & 89.6 & 78.0 & 64.2 & 47.9 & 91.5 & 89.4 & 76.6 & 70.0 & 74.7 & 76.4 & 79.3 & 72.3 \\
\rowcolor[cmyk]{0.05,0.02,0,0}\hspace{2mm}+\ourmethod{}-15M & 336 & 94.7 & 85.1 & 69.7 & 58.5 & 98.8 & \textbf{99.2} & 93.5 & 96.2 & 89.2 & 91.3 & 88.0 & 84.8 \\
\rowcolor[cmyk]{0.05,0.02,0,0}\hspace{2mm}+\ourmethod{}-60M & 336 & \textbf{95.9} & \textbf{85.4} & \textbf{72.4} & \textbf{58.8} & 98.8 & \textbf{99.2} & 95.5 & 97.3 & \textbf{90.7} & \textbf{92.6} & 91.0 & \textbf{86.8} \\
\hline
SigLIP~\cite{zhai2023sigmoid} & 384 & 93.7 & 81.4 & 72.0 & 53.9 & 90.1 & 90.4 & 76.3 & 74.4 & 77.8 & 79.5 & 82.0 & 75.9 \\
\hline
\multicolumn{14}{l}{\textbf{SO/14 (428M)}} \\
\hline
SigLIP~\cite{zhai2023sigmoid} & 224 & 91.0 & 75.2 & 69.8 & 51.8 & 56.1 & 50.0 & 31.5 & 29.1 & 42.3 & 44.8 & 58.1 & 50.2 \\
SigLIP2~\cite{tschannen2025siglip} & 224 & 93.9 & 82.9 & 72.0 & 55.5 & 90.2 & 87.2 & 75.7 & 74.5 & 77.0 & 78.9 & 81.8 & 75.8 \\
\rowcolor[cmyk]{0.05,0.02,0,0}\hspace{2mm}+\ourmethod{}-60M & 224 & \textbf{95.2} & \textbf{84.6} & \textbf{73.7} & \textbf{57.6} & \textbf{99.4} & \textbf{99.4} & \textbf{96.2} & \textbf{97.3} & \textbf{91.5} & \textbf{92.9} & \textbf{91.2} & \textbf{86.4} \\
SigLIP~\cite{zhai2023sigmoid} & 384 & 94.3 & 83.0 & 72.4 & 54.3 & 91.6 & 89.4 & 74.5 & 73.6 & 74.5 & 75.8 & 81.5 & 75.2 \\
\hline
\multicolumn{14}{l}{\textbf{InternVL (6B)}} \\
\hline
InternVL~\cite{chen2024internvl} & 224 & 94.2 & 81.6 & 69.4 & 53.2 & 92.4 & 90.5 & 79.6 & 77.7 & 79.2 & 81.6 & 83.0 & 76.9 \\
\hline
\multicolumn{14}{l}{\textbf{VLM2Vec (7B)}} \\
\hline
VLM2Vec*~\cite{Jiang2024VLM2VecTV} & 1344 &91.6 & 79.8 & 61.3 & 51.5 & 93.9 & 91.0 & 90.9 & 92.0 & 80.5 & 86.0 & 83.6 & 80.1 \\
\hline
\end{tabular}
\caption{Systematic comparison of model performance on multiple datasets. * means reproduced by ourselves. }
\label{tab:sys_compare}
\end{table*}

\section{Methods}

\subsection{Stage 1: LLM Caption Contrastive Fine-tuning}
Traditionally, an LLM’s output layer functions as a classification head that produces discrete text tokens. Recent studies—such as LLM2Vec \cite{llm2vec} and NV-EMBED-v2 \cite{lee2024nv}—have shown that, with additional design tweaks, an LLM can be repurposed as a text-embedding model. Yet these efforts do not account for CLIP’s cross-modal pre-training regime. Here, we advance the discussion from three fronts—\textit{model architecture}, \textit{training methods} and \textit{training data}—explicitly targeting the seamless integration of LLMs with CLIP.

\textbf{Model Architecture:}
\begin{itemize}
    \item \textbf{Sentence Token Representation:} We map the output layer features of the LLM to represent whole sentences. We considered two strategies: using the [EOS] token, which can attend to the entire sentence during pretraining, or employing average pooling across all output tokens. Empirically, average pooling performed better and is our default setting.
    \item \textbf{Bidirectional Attention:} Given that the generative capability is unnecessary for our encoder-focused application, we remove the attention mask from the LLM, enabling bidirectional textual relationship modeling for enhanced comprehension.
    \item \textbf{Fine-tuning Parameters:} To efficiently activate stronger textual comprehension capabilities within the output features of LLMs, we apply parameter-efficient fine-tuning using Low-Rank Adaptation (LoRA).
    \item \textbf{Adaptor Design:} Adding an adaptor following the LLM can potentially enhance feature separability. We experimented with three adaptor variants: (1) a Transformer-based adaptor composed of several latent cross-attention transformer layers following NV-EMBED-v2, (2) a Linear adaptor consisting of multiple linear layers directly attached after extracting sentence tokens from LLM outputs and (3) without using adaptor. 
\end{itemize}

\textbf{Training Methods:}
\begin{itemize}
    \item \textbf{Masked Next Token Prediction (MNTP):} Inspired by masked language modeling in BERT, MNTP improves feature quality by masking and predicting specific tokens. Unlike BERT, predictions are always made at the position preceding the masked token, aligning with LLM's next-token prediction convention. While effective in pure text tasks (as shown by LLM2Vec), we find out that MNTP alone or combined with contrastive learning underperformed relative to direct contrastive fine-tuning in multimodal scenarios.
    \item \textbf{Textual Contrastive Learning:} Following SimCSE, we employ textual contrastive learning to enhance feature separability, pulling positive samples closer and pushing negative samples apart in feature space. Two variants are possible: unsupervised (using dropout augmentation) and supervised (leveraging semantic pairs from annotated data). We default to supervised contrastive learning due to superior performance. Specifically, we generate positive pairs using two distinct captions from the same image, framed by a system prompt: "Given a caption, retrieve a similar relevant caption."
\end{itemize}

\textbf{Training Data:}

To leverage LLM embeddings for \emph{CLIP fine-tuning}, we fine-tune the LLM on caption corpora extracted from image–text datasets, aligning its embedding space with downstream multimodal tasks. Experiments use the Dreamlip~\cite{dreamlip} captions, which provide multiple captions per image, suitable for supervised textual contrastive learning. To preserve general language understanding, we also incorporate 1.5M pure-text pairs from Echo Embeddings~\cite{springer2024repetition} into a mixed contrastive training setup.

\subsection{Stage 2:~\ourmethod{} Post Fine-tuning}
\label{sec:method_stage2}
We aim to perform post fine-tuning on pretrained CLIP since we believe the fundamental visual-language mapping has been well-established through CLIP’s extensive pretraining. By introducing an LLM, we expect to further enrich this representation space, potentially improving CLIP’s capabilities through a highly cost-effective fine-tuning step. CLIP inherently consists of a Vision Encoder (typically a ViT) and a relatively small Text Encoder (roughly one-third the parameters of the Vision Encoder), structured as a compact autoregressive model. Integrating a caption-contrastive fine-tuned LLM as an additional text encoder and finding an effective approach to enhance CLIP performance is critical. Here, we also elaborate on \textit{model architecture}, \textit{training methods} and \textit{training data}.

\textbf{Model Architecture: }
Training CLIP typically requires large-scale datasets and substantial batch sizes, resulting in considerable computational costs. Therefore, we employ efficient Parameter-Efficient Fine-Tuning (PEFT) methods for this stage. By default, we enable full gradients for the visual encoder to facilitate learning from LLM knowledge. Given the large size of the LLM, we compare two strategies: (1) fine-tuning the LLM with LoRA, and (2) freezing LLM gradients and appending previously introduced Transformer or Linear adaptors after its output layer as learnable modules to obtain sentence token features. Method (2) offers clear advantages: it completely eliminates LLM gradient updates, significantly reducing GPU memory usage. Moreover, since the adaptor is placed after the LLM, text features can be precomputed and stored via offline inference, reducing inference overhead from multiple epochs to a single pass. As a result, this approach avoids loading the LLM into GPU memory during training, accelerating experimentation and reducing training time and memory footprint. By default, we adopt method (2) with an adaptor of four linear blocks. Detailed efficiency analysis is provided in \Cref{sec:exp_stage_2}.

\textbf{Training Methods:}
After introducing an LLM into CLIP, we are left with two text encoders.  How should they be used most effectively?
We tested several approaches: (a) Removing the original CLIP Text Encoder and using LLM with CLIP Vision Encoder directly for cross-modal contrastive learning. (b) Retaining both LLM and original Text Encoder, applying separate CLIP losses to both encoders. (c) Adding an additional contrastive learning task between LLM and the original Text Encoder to method (b). (d) Concatenating LLM and Text Encoder output features in method (b), then performing contrastive learning with CLIP Vision Encoder.
Our extensive experiments led us to adopt method (a) for its simplicity and effectiveness.

\textbf{Training Data:}
With the integration of LLM, our method inherently excels at processing dense and detailed image captions. Datasets such as DreamLIP and Recaption, which leverage Share-Captioner~\cite{chen2023sharegpt4v}, InstructBLIP~\cite{instructblip_paper} and LLaVA-1.5~\cite{liu2023improvedllava} to generate multiple dense captions per image, can be effectively utilized. We explored multiple caption scenarios, employing multiple positive caption examples during contrastive training and blending real and MLLM-generated dense captions. Ultimately, we adopted a strategy blending real captions and dense MLLM captions at a 50\% ratio.

\section{Experiments}

\subsection{Default Experimental Settings}
\label{exp:setting}

\noindent\textbf{Default Settings for Stage-1} We choose to use LLaMA 3.1 8B as the LLM integrated into CLIP pre-training, removing the causal attention mask to enable bidirectional attention. No additional adapters are used in caption contrastive finetune, and average pooling is employed to obtain the sentence token embedding. MNTP is not used by default.

We utilize 30M DreamLIP caption data to perform caption contrastive fine-tuning on the LLM. Randomly sampling any two captions of an image from DreamLIP as positive pairs, we conduct supervised SimCSE training. 

\textbf{Default Settings for Stage-2} After caption contrastive fine-tuning, we replace the original CLIP text encoder with the LLaMA-3.1-8B to perform cross-modal contrastive learning fine-tuning with the vision encoder. By default, we open the ViT gradients and freeze the LLM gradients, then add an adapter layer composed of four linear layers after the LLM. The design of the Adaptor is the same as FuseMix~\cite{fusemix}, using a simple inverted bottleneck MLP architecture. By default, we use 15M DreamLIP annotated subsets of CC 3M and CC 12M. For the 60M data setting, we use DreamLIP re-annotated CC 3M and CC 12M, YFCC 15M, and a 30M subset of the LAION dataset. For the 3M setting, we only use the CC 3M subset.

\subsection{System Comparison}
\label{exp:msys}

In the system comparison, we used the default settings from ~\Cref{exp:setting}. However, we also conducted experiments on different pretrained SOTA CLIP models to verify that applying LLM2CLIP post training can significantly enhance their performance.

\textbf{Zero-Shot English Text-Image Retrieval}
For short-text retrieval, we used the MSCOCO~\cite{MSCOCO} 5K test set and the Flickr~\cite{flickr} 1K test set. For long-text retrieval, we employed a 1K subset of ShareGPT4V~\cite{chen2023sharegpt4v} and the Urban1K~\cite{zhang2024long} dataset from LongCLIP~\cite{zhang2024long}, along with the DOCCI~\cite{onoe2024doccidescriptionsconnectedcontrasting} dataset.

The ShareGPT4V-1M dataset consists of captions generated using GPT-4V and ShareCaptioner, covering images from LAION, CC, SBU~\cite{NIPS2011_5dd9db5e}, and MS COCO. Urban1K includes captions for 1,000 urban scene images, each richly annotated with detailed descriptions. DOCCI contains 1.5K high-resolution images with human-annotated captions and was used for retrieval evaluation. To facilitate observation, we additionally report the average Top-1 image-to-text (I2T) and text-to-image (T2I) accuracy across these five datasets.

\textbf{\ourmethod{} Improves CLIP for Long- and Short-Text Retrieval.} In ~\Cref{tab:sys_compare}, we compared OpenAI’s CLIP, EVA02, and SigLIP2, and conducted experiments using three types of visual models: ViT-B/16, ViT-L/14, and SoViT/14. Regardless of the model size or resolution shown in the table, \ourmethod{} yields a significant performance boost for both CLIP and EVA02. For example, under EVA02 at 224 resolution, the LLM2CLIP-60M design achieves an average Top-1 retrieval accuracy improvement of +12 and +14.8. Even for SigLIP2—which was pretrained on 40B data—our method still brings an average performance improvement of +9.4 and +10.6; even on the short-caption datasets Flickr and COCO, where SigLIP2 is already very familiar, improvements of +1 and +1.9 are observed. Comparatively, the improvements for long captions are even larger, as the inherent large window of LLMs fully leverages their advantage in understanding long and complex texts. Notably, \ourmethod{} even surpasses \textit{InternVL}—which contains 6 B trainable parameters—and \textit{VLM2Vec}, whose pipeline fine-tunes the entire LLM with LoRA.


\begin{table}[h]
\centering
\footnotesize
\begin{tabular}{@{}l|cc|cc|cc@{}}
\hline
\multirow{2}{*}{\textbf{Models}} &
\multicolumn{2}{c|}{\textbf{Flickr-CN}} &
\multicolumn{2}{c|}{\textbf{COCO-CN}} &
\multicolumn{2}{c}{\textbf{XM3600}} \\
\cline{2-7}
& \textbf{I2T} & \textbf{T2I} & \textbf{I2T} & \textbf{T2I} & \textbf{I2T} & \textbf{T2I} \\
\hline
CN-CLIP & 80.2 & 68.0 & 63.4 & 64.0 & -- & -- \\
EVA-L-224 & 4.4 & 0.9 & 2.6 & 1.0 & 14.0 & 8.0 \\
\rowcolor[cmyk]{0.05,0.02,0,0}\hspace{2mm}+\ourmethod{} & \textbf{90.6} & 75.6 & \textbf{72.0} & 70.1 & 68.3 & 56.0 \\
SigLIP2 & 79.2 & 56.9 & 55.3 & 51.7 & 59.7 & 48.2 \\
\rowcolor[cmyk]{0.05,0.02,0,0}\hspace{2mm}+\ourmethod{} & 90.0 & \textbf{76.1} & 70.8 & \textbf{70.2} & \textbf{69.1} & \textbf{56.3} \\
\hline
\end{tabular}
\caption{Multi-lingua retrieval results on different datasets.}
\label{tab:flickr_coco_cn_results}
\end{table}

\begin{table}[h]
\centering
\footnotesize
\begin{tabular}{@{}l|ccc@{}}
\toprule
\multirow{2}{*}{\textbf{Methods}} & \multicolumn{3}{c}{\textbf{Imagenet}} \\
 & \textbf{0-shot*} & \textbf{0-shot} & \textbf{Linear} \\
\midrule
\textbf{CLIP L/14-336} & \textbf{74.9} & \textbf{76.6} & 84.8 \\
\rowcolor[cmyk]{0.05,0.02,0,0}\hspace{2mm}+\ourmethod{} & 74.6 & 75.8 & \textbf{85.2} \\
\textbf{CLIP L/14} & \textbf{73.7} & \textbf{75.5} & 83.9 \\
\rowcolor[cmyk]{0.05,0.02,0,0}\hspace{2mm}+\ourmethod{} & 73.2 & 74.3 & \textbf{84.4} \\
\bottomrule
\end{tabular}
\caption{Zero-shot classification and linear probe performance on ImageNet. *0-shot uses the class template \emph{`a photo of the \{classname\}}' only.}
\label{tab:imagenet}
\end{table}

\begin{table}[h]
\centering
\scriptsize
\setlength{\tabcolsep}{0.9mm}
\begin{tabular}{lcccccccc}
\toprule
\multirow{3}{*}{Method} & \multicolumn{4}{c}{Zero-shot Seg. mIOU} & \multicolumn{3}{c}{OV-COCO Det.} & \multicolumn{1}{c}{COCO val2017} \\
\cmidrule(lr){2-5} \cmidrule(lr){6-8} \cmidrule(lr){9-9}
                        & COCO-S & ADE & VOC & City & Novel & Base & All & AP$^{bb}$/AP$^{seg}$ \\
\midrule
EVA02                   & 12.9 & 11.5 & 21.0 & 13.5 & 24.7 & 53.6 & 46.0 & 45.0/38.2 \\
\rowcolor[cmyk]{0.05,0.02,0,0}
+LLM2CLIP               & \textbf{15.3} & \textbf{15.8} & \textbf{29.1} & \textbf{20.1} & \textbf{28.9} & \textbf{54.7} & \textbf{48.0} & \textbf{45.6}/\textbf{38.7} \\
\bottomrule
\end{tabular}
\caption{Zero-shot/supervised segmentation, open-vocabulary detection benchmarks. COCO val2017 is supervised results using CLIP's visual encoder. }
\label{tab:compressed_vision_results}
\end{table}

\begin{table*}[ht]
\centering
\scriptsize 
\begin{tabular}{@{}l|ccccc|ccc|ccccc|ccc@{}}
\toprule
\multirow{2}{*}{\textbf{Model}} & \multicolumn{5}{c|}{\textbf{VQA}} & \multicolumn{3}{c|}{\textbf{Pope}} & \multicolumn{5}{c|}{\textbf{MM}} & \multicolumn{3}{c}{\textbf{Seed}} \\
 & V2 & GQA & Vz & SQA & TV & R & A & P & MME & MB & MC & LB & MV & All & I & V \\
\midrule
\textcolor[cmyk]{0,0,0,0.5}{\textbf{Llava (Paper)}} & \textcolor[cmyk]{0,0,0,0.5}{78.5} & \textcolor[cmyk]{0,0,0,0.5}{62.0} & \textcolor[cmyk]{0,0,0,0.5}{50.0} & \textcolor[cmyk]{0,0,0,0.5}{66.8} & \textcolor[cmyk]{0,0,0,0.5}{58.2} & \textcolor[cmyk]{0,0,0,0.5}{87.3} & \textcolor[cmyk]{0,0,0,0.5}{86.1} & \textcolor[cmyk]{0,0,0,0.5}{84.2} & \textcolor[cmyk]{0,0,0,0.5}{1510.7} & \textcolor[cmyk]{0,0,0,0.5}{64.3} & \textcolor[cmyk]{0,0,0,0.5}{58.3} & \textcolor[cmyk]{0,0,0,0.5}{65.4} & \textcolor[cmyk]{0,0,0,0.5}{31.1} & \textcolor[cmyk]{0,0,0,0.5}{58.6} & \textcolor[cmyk]{0,0,0,0.5}{66.1} & \textcolor[cmyk]{0,0,0,0.5}{37.3} \\
\textbf{Llava (Rep.)} & 79.04 & 62.86 & 50.57 & 67.97 & 57.48 & 87.7 & \textbf{84.85} & 86.3 & 1476.69 & 66.66 & 60.39 & 58.0 & 34.3 & 59.86 & 66.95 & \textbf{39.71} \\
\rowcolor[cmyk]{0.05,0.02,0,0}\hspace{2mm}+\ourmethod{} & \textbf{79.80} & \textbf{63.15} & \textbf{52.37} & \textbf{69.92} & \textbf{58.35} & \textbf{88.55} & 82.76 & \textbf{87.75} & \textbf{1505.82} & \textbf{68.29} & \textbf{60.40} & \textbf{62.7} & \textbf{34.8} & \textbf{60.96} & \textbf{68.80} & 38.96 \\
\bottomrule
\end{tabular}
\caption{Performance of Llava 1.5 benchmarks. For +~\ourmethod{} we replace Llava's CLIP Vit-L/14 with our finetuned version. }
\label{tab:model_performance_comparison}
\end{table*}

\textbf{Fine-tuning Data Volume Analysis. } 
In~\Cref{tab:sys_compare}, we experimented with 3M, 15M, and 60M training data on CLIP and EVA02 using ViT-L/14. More data consistently improves retrieval performance for both long and short text tasks at 224 and 336 resolutions. With 3M data, EVA02 shows noticeable improvement in long text retrieval but minimal gain in short text retrieval, suggesting the model first compensates for disrupting the original cross-modal space before achieving broader improvements.

\textbf{Zero-Shot Multilingual Text-Image Retrieval}
As in~\Cref{tab:flickr_coco_cn_results}, We conducted cross-lingual retrieval experiments where CLIP vision encoders were trained exclusively on English text. We successfully endowed EVA02 with multilingual capabilities, which it previously lacked, and significantly enhanced the already robust multilingual performance of SigLIP2.

\textbf{Zero-shot \& Linear Probe ImageNet Classification}  
We evaluate our model on ImageNet using two standard protocols:  
(1) \textbf{Zero-shot classification}, where we report accuracy using the prompt template ``a photo of the \{classname\}'', following the CLIP methodology. We also evaluate the average performance over 80 handcrafted prompt variants, as commonly practiced in prior work.  
(2) \textbf{Linear probing}, where we freeze the visual encoder and train a linear classifier using fixed hyperparameters across all experiments (batch size 1024, learning rate 0.1, momentum 0.9, weight decay 0, 50 epochs with SGD).

As shown in \Cref{tab:imagenet}, we conduct experiments using CLIP ViT-L/14 at both 224 and 336 resolutions.  
We observe that after applying \ourmethod{}, zero-shot performance on ImageNet shows a modest drop.  
However, the accuracy from linear probing \emph{improves} over the original CLIP baseline.  
This suggests that while the quality of the learned \emph{visual features} remains strong (as linear probing only tests CLIP visual encoder’s representation power using a supervised head), the overall \emph{multimodal alignment} in the shared space may slightly deteriorate since zero-shot classification drops, especially for fine-grained category separation.

We hypothesize that the drop in zero-shot performance could be mitigated by increasing the amount of LLM2CLIP fine-tuning data, which remains to be verified in future work.  
Nonetheless, this trade-off echoes a common pattern observed in recent CLIP fine-tuning studies, such as LongCLIP~\cite{zhang2024long} and CLIP-MoE~\cite{zhang2024clipmoe}, where performance on head classes improves while long-tail categories suffer due to distribution imbalance—particularly under limited-scale fine-tuning. The contrast with retrieval results is instructive: retrieval benefits greatly because its vocabulary is broad and frequent, whereas classification demands uniform discriminability across all (often obscure) class nouns.

\textbf{Zero-Shot / Supervised Segmentation and Object Detection}
As our linear-probe results already suggest, the visual features produced by the encoder become noticeably stronger after LLM2CLIP training.~\Cref{tab:compressed_vision_results} further confirms this on both zero-shot and fully supervised object detection and segmentation benchmarks. In the zero-shot setting—which relies on the text encoder—LLM2CLIP improves CLIP’s multimodal representations; in the supervised setting—where only the vision encoder is finetuned—it also enhances the purely visual, low-level feature extractor. We conjecture that these gains stem from the LLM’s ability to parse dense captions more accurately, especially spatial terms and object-to-object relations, and to inject that knowledge into the joint vision–language space.

\textbf{Multimodal Large Language Models Performance}
Following LLAVA1.5~\cite{liu2023llava}, we used OpenAI’s CLIP-ViT-L-336 encoder with a simple MLP head to connect to Vicuna-7B. Pretraining included 558K image-caption pairs and 665K visual instruction samples. We fine-tuned the LLAVA1.5 encoder using \ourmethod{} to assess improvements in visual feature extraction. As shown in~\Cref{tab:model_performance_comparison}, \ourmethod{} fine-tuning enhanced multimodal model performance on over 87.5\% of benchmarks, with minor losses in only two tasks. This highlights the potential of \ourmethod{} for improving CLIP visual encoder's abilities for complex image reasoning and understanding.


\begin{table}[h]
\centering
\scriptsize  
\setlength{\tabcolsep}{1mm}
\begin{tabular}{@{}l|cc@{}}
\hline
\multirow{2}{*}{\textbf{Stage1}} & \multicolumn{2}{c}{\textbf{Avg}} \\
 & \textbf{I2T} & \textbf{T2I} \\
\hline
\rowcolor[cmyk]{0.05,0.02,0,0} Lora, AvgPool, Bidirectional, Supervise Simcse & \textbf{80.4} & \textbf{77.9} \\
Lora, AvgPool, Bidirectional, \textcolor[cmyk]{0,1,1,0}{Un-supervise Simcse} & 59.2 & 57.7 \\
Lora, AvgPool, Bidirectional, \textcolor[cmyk]{0,1,1,0}{MNTP} & 70.1 & 67.0 \\
Lora, AvgPool, Bidirectional, \textcolor[cmyk]{0,1,1,0}{MNTP}, Supervise Simcse & 79.7 & 77.2 \\
Lora, AvgPool, \textcolor[cmyk]{0,1,1,0}{Casual}, Supervise Simcse & 80.0 & 77.5 \\
Lora, \textcolor[cmyk]{0,1,1,0}{EOS}, Bidirectional, Supervise Simcse & 80.0 & 77.3 \\
\textcolor[cmyk]{0,1,1,0}{Frozen}, \textcolor[cmyk]{0,1,1,0}{Linear Adaptor}, AvgPool, Bidirectional, Supervise Simcse & 74.1 & 71.3 \\
\hline
\end{tabular}
\caption{Ablation study on the training methods of LLM caption contrastive finetuning in Stage 1. \textcolor{red}{red} text indicates content that differs from the default setting.}
\label{tab:stage1_methods}
\end{table}

\begin{table}[h]
\centering
\footnotesize
\begin{tabular}{@{}l|cc@{}}
\hline
\multirow{2}{*}{\textbf{Method}} & \multicolumn{2}{c}{\textbf{Avg}} \\
 & \textbf{I2T} & \textbf{T2I} \\
\hline
CLIP & 74.4 & 72.0 \\
Directly Finetune (50\%) & 74.5 & 72.3 \\
\hline
bge-en-icl & 78.9 & 78.2 \\
LLM2Vec-Llama-3-8B & 81.4 & 80.2 \\
NV-Embed-v2 & 81.4 & 79.9 \\
\hline
VLM2Vec & 78.2 & 77.1 \\
\hline
bge-m3-XLM-R & 65.0 & 63.6 \\
jina-v3-XLM-R & 73.6 & 71.0 \\
e5 (XLM-R) & 74.0 & 71.7 \\
\hline
Qwen2.5-0.5B-CC & 75.6 & 73.0 \\
Llama-3.2-1B-CC & 80.4 & 77.9 \\
Llama-3-8B-CC & 83.4 & 80.9 \\
DeepSeek-R1-Distill-Llama-8B-CC & 83.5 & 80.5 \\
Llama-3.1-8B-CC & \textbf{84.8} & \textbf{81.0} \\
\rowcolor[cmyk]{0.1,0.2,0,0}\textcolor[cmyk]{0,1,1,0}{Llama3.1-8B} & 66.5 & 62.5 \\
\hline
\end{tabular}
\caption{Ablation for using different text encoder. "-CC" indicates encoders with caption contrastive fine-tuning.}
\label{tab:llm_encoders}
\end{table}

\subsection{Ablation Study: Stage-1 Caption Contrastive Fine-tuning}
\label{exp:ccfinetune}
For simplicity, we select Llama 3.1 1B as the default LLM to investigate different design choices for caption contrastive fine-tuning in this section. We use a subset of the CC 3M dataset for caption contrastive fine-tuning. Additionally, we explore augmenting the training data with the Wikitext-103 dataset~\cite{Merity_Xiong_Bradbury_Socher_2016} for MNTP training and the E5 dataset~\cite{springer2024repetition} as supplementary pure-text data for caption contrastive fine-tuning.

\begin{table}[h]
\centering
\footnotesize 
\begin{tabular}{@{}l c | cc@{}}
\hline
\multirow{2}{*}{\textbf{Stage1}} & \multirow{2}{*}{\textbf{Stage2}} & \multicolumn{2}{c}{\textbf{Avg}} \\
 &  & \textbf{I2T} & \textbf{T2I} \\
\hline
-- & -- & 78.3 & 75.5 \\
-- & Linear($\times1$) & 79.2 & 76.7 \\
-- & Linear($\times2$) & 80.1 & 76.8 \\
\rowcolor[cmyk]{0.05,0.02,0,0}-- & Linear($\times4$) & 80.4 & \textbf{77.9} \\
Linear($\times4$) & Linear($\times4$) & \textbf{80.5} & 77.7 \\
-- & Transformer($\times1$) & 80.2 & 77.3 \\
Transformer($\times1$) & Transformer($\times1$) & \textbf{80.5} & 77.3 \\
\hline
\end{tabular}
\caption{Ablation experiments for adaptor design on Stage 1 caption contrastive finetune and Stage 2 \ourmethod{} post training. The 4-layer Linear Adaptor has 67.1M parameters, while the single layer Transformer Adaptor has 67.6M. }
\label{tab:adaptor_analysis}
\end{table}

\textbf{Architecture Design Ablation.}
We examine various adaptor architecture in~\Cref{tab:adaptor_analysis}. Comparing rows 4 and 5, as well as rows 6 and 7, we observe that neither Linear nor Transformer adaptors significantly impact the performance of subsequent CLIP fine-tuning. Therefore, we default to using \textit{no adaptor} in Stage 1.

\begin{table}[ht]
\centering
\scriptsize     
\setlength{\tabcolsep}{0.9mm}
\begin{tabular}{l c | c c | c c}
\hline
\multicolumn{2}{c|}{\multirow{2}{*}{\textbf{Methods}}} & \multirow{2}{*}{\textbf{Training Loss}} & \multirow{2}{*}{\begin{tabular}[c]{@{}c@{}}Testing Text\\Encoder\end{tabular}} & \multicolumn{2}{c}{\textbf{Average}} \\
 &  &  &  & \textbf{I2T} & \textbf{T2I} \\
\hline
CLIP              &               & CL(CLIP-T, CLIP-V) & CLIP-T & 74.4 & 72.0 \\
Directly Finetune &               & CL(CLIP-T, CLIP-V) & CLIP-T & 74.5 & 72.3 \\
\hline
\multirow{10}{*}{\begin{tabular}[c]{@{}l@{}}+\ourmethod{}\end{tabular}} 
                  & {a)} & CL(LLM, CLIP)      & LLM    & 83.9 & 82.1 \\
\cline{3-6}
                  & \multirow{2}{*}{b)} & \multirow{2}{*}{\begin{tabular}[c]{@{}c@{}}CL(LLM, CLIP-V)+\\CL(CLIP-T, CLIP-V)\end{tabular}} & CLIP-T & 74.4 & 72.0 \\
                  &                   &                                          & LLM    & 83.6 & 81.8 \\
\cline{3-6}
                  & \multirow{3}{*}{c)} & \multirow{3}{*}{\begin{tabular}[c]{@{}c@{}}CL(LLM, CLIP-V)\\+CL(CLIP-T, CLIP-V)\\+CL(CLIP-T, LLM)\end{tabular}} & CLIP-T & 74.0 & 72.1 \\[3ex]
                  &                   &                                          & LLM    & 83.7 & 81.4 \\
\cline{3-6}
                  & \multirow{3}{*}{d)} & \multirow{3}{*}{\begin{tabular}[c]{@{}c@{}}CL(LLM, CLIP-V)\\+CL(CLIP-T, CLIP-V)+\\CL(Cat(CLIP-T, LLM), CLIP-V)\end{tabular}} & CLIP-T & 74.8 & 71.3 \\
                  &                   &                                          & LLM    & 83.8 & 82.3 \\
                  &                   &                                          & \begin{tabular}[c]{@{}c@{}}Cat(CLIP-T,\\LLM)\end{tabular} & \textbf{84.7} & \textbf{82.8} \\
\cline{3-6}
\hline
\end{tabular}
\caption{\ourmethod{} Training Method Analysis. We experimented with various possibilities for fine-tuning between the LLM and the pretrained CLIP model's Vision Encoder (CLIP-V) and Text Encoder (CLIP-T). "CL" denotes the contrastive learning loss, and "Cat" means concatenation. Items a--d correspond to the a--d in~\Cref{sec:method_stage2}. }
\label{tab:stage-2-methods} 
\end{table}
\textbf{Training Method Ablation.}
We conduct an ablation study of Stage-1 LLM caption contrastive fine-tuning methods in ~\Cref{tab:stage1_methods}, observing the following key insights:
1. Freezing LLM gradients and relying solely on adaptors yields poor performance, indicating that enabling LoRA is essential.
2. SimCSE is identified as the most critical loss function. MNTP does not significantly enhance SimCSE, and relying solely on MNTP without SimCSE severely deteriorates performance.
3. Supervised SimCSE substantially outperforms unsupervised SimCSE.
4. The performance difference between causal and bidirectional attention is negligible.

\textbf{Ablation across different LLM backbones.}  
Table~\ref{tab:llm_encoders} compares a variety of text-embedding models plugged into our pipeline.  
Our approach delivers substantial gains over all contenders.  
The line highlighted in \textcolor{red}{red} shows that \emph{plain} Llama\,3.1-8B—without our caption-contrastive (CC) fine-tuning—performs very poorly and even harms the original CLIP, underscoring the necessity of the CC stage.  
We also evaluated several state-of-the-art embedding models, including \texttt{bge}, \texttt{jina-v3}, and the VLM2Vec embedding obtained from a multimodal LLM fine-tune.  
In every case, the \ourmethod{} variant still achieves the best results, confirming the effectiveness of our training recipe.

\subsection{Ablation Study: Stage-2~\ourmethod{}}
\label{sec:exp_stage_2}

\noindent\textbf{Adaptor Design.}
As shown in~\Cref{tab:adaptor_analysis}, we explore adaptor structures in the Stage-2 LLM2CLIP cross-modal pre-training. Comparing Linear ($\times4$) and Transformer ($\times1$), their performances are relatively similar, and thus, we select the simpler Linear adaptor structure. Furthermore, in our layer-wise ablation of Linear adaptors, we find that increasing the adaptor size indeed improves performance. Specifically, adaptor performance improves progressively when transitioning from no adaptor to 1, 2 and 4-layer.

\textbf{Training-method ablation.}  
We explored multiple ways to integrate an LLM during CLIP fine-tuning (implementation details appear in~\Cref{tab:stage-2-methods}.  
Comparing experiments~(a)–(b) and~(b)–(c) shows that neither (i) optimising two separate losses nor (ii) explicitly aligning the two text encoders brings any meaningful benefit.  
\emph{This observation motivates our choice to replace, rather than reuse, CLIP’s original text encoder in the main pipeline.}  
Experiment~(d)—which performs contrastive learning on the \emph{concatenated} outputs of both encoders—does deliver a noticeable uplift; for the sake of simplicity, we do not adopt it as our default, but regard it as a promising future work.

\section{Conclusion}  
We have revisited the problem of upgrading CLIP with large-language-model knowledge and proposed a \emph{lightweight} pipeline that first turns an LLM into a discriminative caption-embedding module and then inserts it into CLIP via a cost-effective fine-tuning stage.  
With only \emph{million-scale} training pairs and a compute budget essentially identical to standard CLIP fine-tuning, our method delivers sizeable gains—on top of already \textsc{SOTA} baselines—in image–text retrieval, image classification (zero-shot \& linear probe), zero-shot/supervised detection/segmentation, and even as the visual encoder inside a multimodal LLM. 

\section*{Acknowledgments}
This work is partially supported by the National Natural Science Foundation of China (NSFC Granted No. 62276190).




\bibliography{aaai2026}





\appendix
\renewcommand{\thefigure}{A\arabic{figure}}
\renewcommand{\thetable}{A\arabic{table}}
\setcounter{figure}{0}
\setcounter{table}{0}
\section{Reproducibility Statement}

To ensure the reproducibility of our work, we are committed to making all the training and testing code, along with the datasets used in our experiments, publicly available in our code repository. Additionally, we will open-source our~\ourmethod{}-enhanced versions of SigLip2, EVA02, and OpenAI's CLIP to facilitate applications within the open-source community requiring high-quality embedding models.

We will provide comprehensive documentation alongside the codebase to facilitate straightforward replication of our experiments. All hyperparameter settings, training configurations, and preprocessing steps will be thoroughly documented and made accessible to the research community.

Additionally, we will include detailed environment setup instructions, dependency requirements, and step-by-step execution guidelines to minimize barriers for reproduction. Our commitment to open science ensures that researchers can easily validate, extend, and build upon our contributions.

\section{Models and Checkpoints}
Table~\ref{tab:all_models} lists all the models and checkpoints used in our experiments, ensuring transparency and facilitating reproducibility. It includes both vision and text encoders, with corresponding links to their publicly available implementations or weights.

\section{Experiment Details}
\subsection{Training Setup}
\paragraph{Settings for Stage-1} We choose to use LLaMA 3.1 8B as the LLM integrated into CLIP pre-training, removing the causal attention mask to enable bidirectional attention. No additional adapters are used in caption contrastive finetune, and average pooling is employed to obtain the sentence token embedding. MNTP is not used by default.
Additionally, to ensure the generalizability of LLM textual differentiation capabilities, we incorporated a 1.5M paired pure-text dataset curated by Echo Embeddings~\cite{springer2024repetition} into our contrastive training process.Our models are trained with LoRA ($r=16$, $\alpha=32$), \texttt{bfloat16} quantization, gradient checkpointing, and FlashAttention-2 for memory efficiency. Training uses the AdamW optimizer (learning rate $2\times10^{-4}$, 300-step linear warm-up), sequence length 512, effective batch size 2048, and 1 epoch over 30M(1.5m paired pure-text and captions dataset) samples on 32 NVIDIA A100 GPUs.

\paragraph{Settings for Stage-2} We employed the AdamW optimizer, utilizing a total batch size of 4096, and set the learning rate to 1e-5 with a cosine decay rate of 0.05. For every training step we randomly sampled a single caption from the DreamLIP caption data. All models were trained for 4 epochs by default. The embedding dimension for both modalities was set to 1280.

\subsection{Evaluation}
\subsubsection{Multilingual Retrieval}
We evaluate our models on three cross-modal retrieval benchmarks that support multilingual and cross-lingual evaluation: Flickr-CN, COCO-CN, and XM3600. The results are shown in Table 2 of the main text..

\textbf{Flickr-CN}~\cite{lan2017fluency} and \textbf{COCO-CN}~\cite{li2019coco} are Chinese adaptations of the popular Flickr30K and MS-COCO datasets, respectively. These datasets contain images paired with Chinese captions, enabling evaluation of cross-modal retrieval performance in Chinese. 

\paragraph{Crossmodal-3600(XM-3600)}~\cite{thapliyal2022crossmodal} is a geographically-diverse multilingual dataset comprising 3,600 images annotated with human-generated reference captions in 36 languages~. The images were carefully selected from across the world, covering regions where the 36 languages are natively spoken.

\subsubsection{Open-Vocabulary Semantic Segmentation}
CLIP-like models enable patch-level semantic matching by aligning visual and textual representations, facilitating open-vocabulary semantic segmentation. Following MaskCLIP~\cite{zhou2022extract}, we compute cosine similarity between patch embeddings and sentence embeddings to generate segmentation masks. 

We evaluated our models on four benchmark datasets: ADE20K~\cite{zhou2017scene}, COCO-Stuff164k~\cite{caesar2018coco}, VOC20~\cite{everingham2015pascal}, and Cityscapes~\cite{cordts2016cityscapes} using mean Intersection over Union (mIoU) to measure segmentation accuracy. The results are shown in Table 4 of the main text.

\subsubsection{Open-Vocabulary Object Detection}
To assess the fine-grained localization capabilities of our approach, we integrate~\ourmethod{} as the backbone network for open-vocabulary detection tasks. Following established protocols from prior work~\cite{wu2023clipself}, we adopt a two-stage detection framework for evaluation. We conduct experiments on the OV-COCO benchmark and report box Average Precision (AP) at IoU threshold 0.5 across three categories: base classes (seen during training), novel classes (unseen during training), and all classes combined. The novel class performance particularly demonstrates the model's generalization ability to previously unseen object categories. The strong results on novel classes, as shown in Table 4 of the main paper, demonstrate our method’s effectiveness in recognizing and localizing previously unseen objects, highlighting its robust generalization capabilities for open-vocabulary detection tasks.

\paragraph{Supervised Instance Segmentation.} To evaluate the dense prediction capabilities of our method on supervised instance segmentation tasks, we employ~\ourmethod{} as the backbone for ViTDet~\cite{li2022exploring} within the Mask R-CNN~\cite{he2017mask} framework. The models are trained on the COCO dataset using 4 A100 machines with 2 images per GPU, resulting in a total batch size of 64. We train for 25 epochs using the AdamW optimizer with a learning rate of 0.0001, $\beta_1 = 0.9$, $\beta_2 = 0.999$, and a weight decay of 0.1.

Our approach achieves results with 45.6 mAP for object detection and 38.7 mAP for instance segmentation, as shown in Table 4 of the main text, demonstrating the effectiveness of our cross-modal pre-training in improving dense prediction tasks. These results highlight the model's ability to transfer learned representations to complex downstream tasks requiring both object localization and precise pixel-level segmentation.

\paragraph{Comparisons on General Multimodal Benchmarks.} 
We evaluate~\ourmethod{} (EVA02-L-336) as a visual feature extractor for multimodal large language models, comparing it against the standard CLIP baseline. Our experiments are conducted using LLaVA-v1.5-7B, which originally employs CLIP as its visual encoder. To ensure fair and rigorous comparison, we maintain all parameter configurations identical to the original LLaVA setup and utilize the same training data provided by the LLaVA framework. The comprehensive evaluation results on various multimodal understanding benchmarks((\textit{i.e.}, VQA~\cite{goyal2017making}~\cite{hudson2019gqa}~\cite{gurari2018vizwiz}~\cite{lu2022learn}~\cite{singh2019towards}, POPE~\cite{li2023evaluating},  MMBench~\cite{liu2024mmbench}, SEED~\cite{li2023seed}) are presented in Table 5 of the main text.

\begin{figure}[ht]
    \centering
    \includegraphics[width=1.0\linewidth]{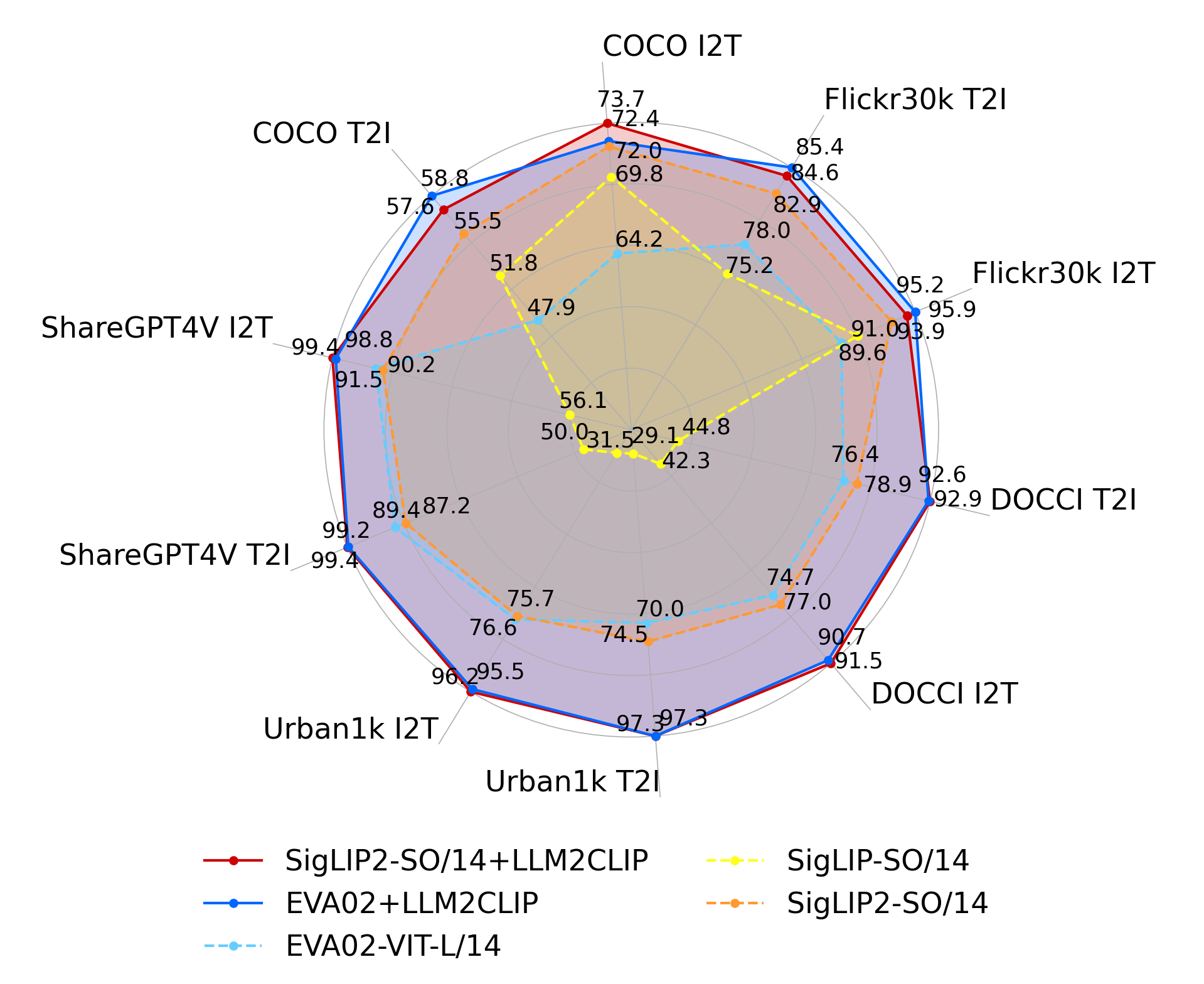}
    \caption{Comprehensive evaluation results on cross-modal retrieval tasks. The radar plot illustrates the performance of different models across multiple benchmarks. Our methods consistently achieve higher scores across all evaluation metrics compared to baseline models.}
    \label{fig:radar}
    
\end{figure}


\section{More Analysis}

\paragraph{Quantitative Analysis of the Original LLM's Feature Separability.}
As shown in~\Cref{tab:crs_lcrs_comparison}, we conducted caption-to-caption retrieval experiments on the COCO 5K test set (each image has five captions; we used the first two captions as positive pairs). We observed that large language models, such as Llama 3 8B, achieve very low accuracy, struggling to distinguish effectively between different captions of the same image. This result highlights the necessity of the embedding-oriented training (caption contrastive fine-tuning) explored in this work for adapting LLMs specifically for the CLIP scenario.

\begin{table}[h]
    \centering
    \resizebox{0.4\linewidth}{!}{ 
        \begin{tabular}{c|c}
            \textbf{Language Model} & \textbf{Top 1 Acc.} \\  
            \shline\hline  
            CLIP-L/14 & 25.2 \\  
            EVA02-L/14 & 27.1 \\  
            \rowcolor{red!20} \textcolor{red}{Llama3-8B} & \textcolor{red}{5.2} \\  
            \rowcolor{red!20} \textcolor{red}{Llama3.2-1B} & \textcolor{red}{5.6} \\  
            Llama3-8B-CC & \textbf{29.5} \\  
            Llama3.2-1B-CC & 29.4 \\  
        \end{tabular}
    }
    \caption{Comparison of top-1 retrieval accuracy for various language models on the MS COCO~\cite{MSCOCO} 5K test set.}
    \label{tab:crs_lcrs_comparison}
\end{table}

\begin{table*}[ht]
\centering
\begin{tabular}{|l|l|}
\hline
\textbf{Model} & \textbf{Link} \\
\hline
\multicolumn{2}{|c|}{\textbf{Vision Encoders}} \\
\hline
ViT-B/16 & \url{https://huggingface.co/openai/clip-vit-base-patch16} \\
ViT-L/14 & \url{https://huggingface.co/openai/clip-vit-large-patch14} \\
ViT-L/14-336 & \url{https://huggingface.co/openai/clip-vit-large-patch14-336} \\
EVA-L-224 & \url{https://huggingface.co/QuanSun/EVA-CLIP/blob/main/EVA02_CLIP_L_psz14_s4B.pt} \\
EVA-L-336 & \url{https://huggingface.co/QuanSun/EVA-CLIP/blob/main/EVA02_CLIP_L_336_psz14_s6B.pt} \\
SigLIP SO/14 & \url{https://huggingface.co/google/siglip-so400m-patch14-224} \\
SigLIP2 SO/14 & \url{https://huggingface.co/google/siglip2-so400m-patch14-224} \\
\hline
\multicolumn{2}{|c|}{\textbf{Text Encoders}} \\
\hline
bge-en-icl & \url{https://huggingface.co/BAAI/bge-en-icl} \\
LLM2Vec-Llama-3-8B & \url{https://huggingface.co/McGill-NLP/LLM2Vec-Meta-Llama-3-8B-Instruct-mntp} \\
NV-Embed-v2 & \url{https://huggingface.co/nvidia/NV-Embed-v2} \\
VLM2Vec & \url{https://huggingface.co/TIGER-Lab/VLM2Vec-LLaVa-Next} \\
bge-m3-XLM-R & \url{https://huggingface.co/BAAI/bge-m3} \\
jina-v3-XLM-R & \url{https://huggingface.co/jinaai/jina-embeddings-v3} \\
e5 & \url{https://huggingface.co/intfloat/multilingual-e5-large} \\
Qwen2.5-0.5B & \url{https://huggingface.co/Qwen/Qwen2.5-0.5B} \\
Llama-3.2-1B & \url{https://huggingface.co/meta-llama/Llama-3.2-1B} \\
Llama-3-8B & \url{https://huggingface.co/meta-llama/Meta-Llama-3-8B} \\
DeepSeek-R1-Distill-Llama-8B & \url{https://huggingface.co/deepseek-ai/DeepSeek-R1-Distill-Llama-8B} \\
Llama-3.1-8B & \url{https://huggingface.co/meta-llama/Llama-3.1-8B} \\
\hline
\end{tabular}
\caption{Models and Checkpoints used in our experiments, including both vision and text encoders.}
\label{tab:all_models}
\end{table*}

    


\begin{figure*}[!h]
    \centering
    \begin{subfigure}{0.48\linewidth}
        \centering
        \includegraphics[width=\linewidth]{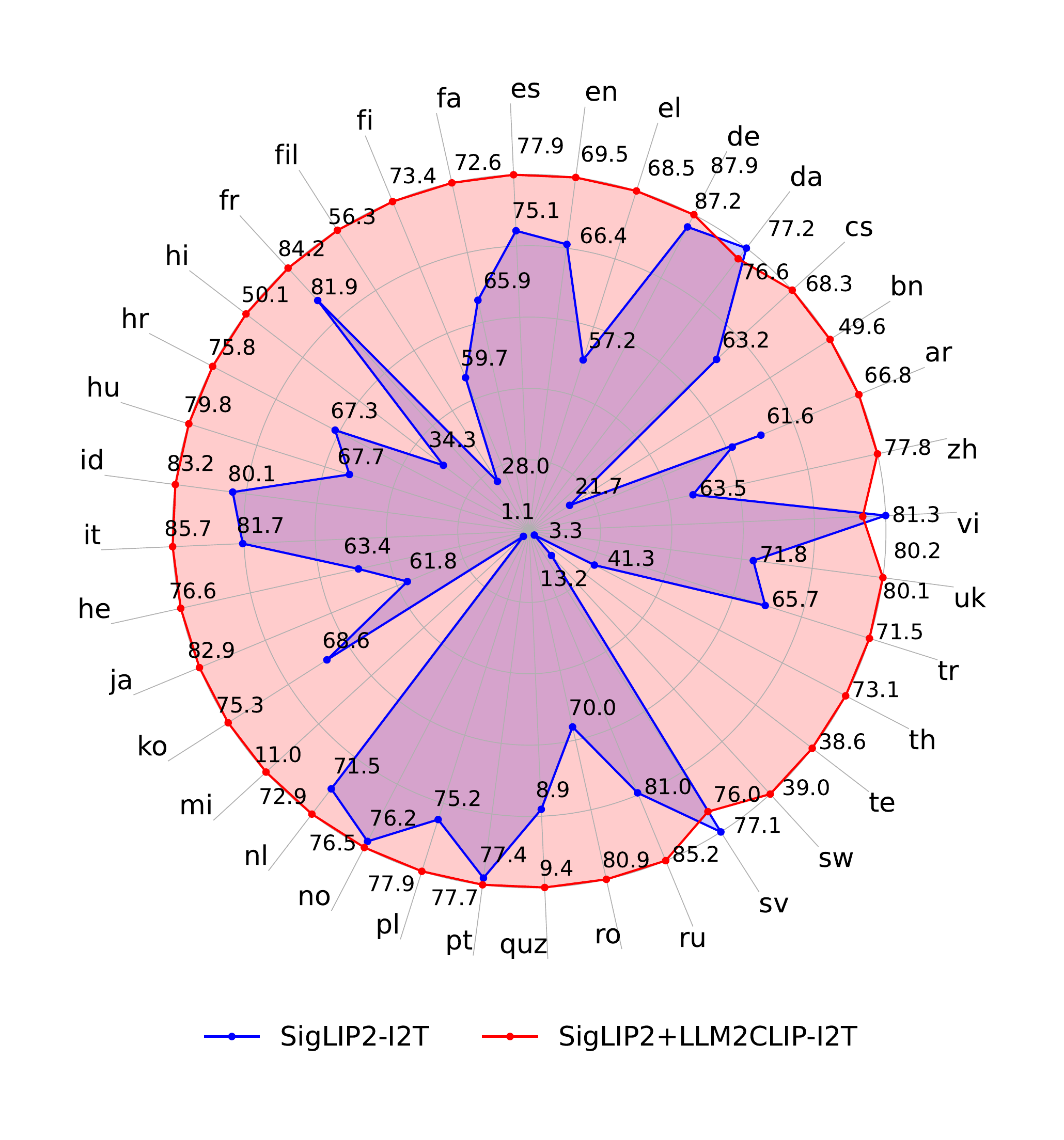}
        \caption{Multilingual zero-shot image-to-text retrieval (Recall@1)}
        \label{fig:xm_i2t}
    \end{subfigure}
    \hfill
    \begin{subfigure}{0.48\linewidth}
        \centering
        \includegraphics[width=\linewidth]{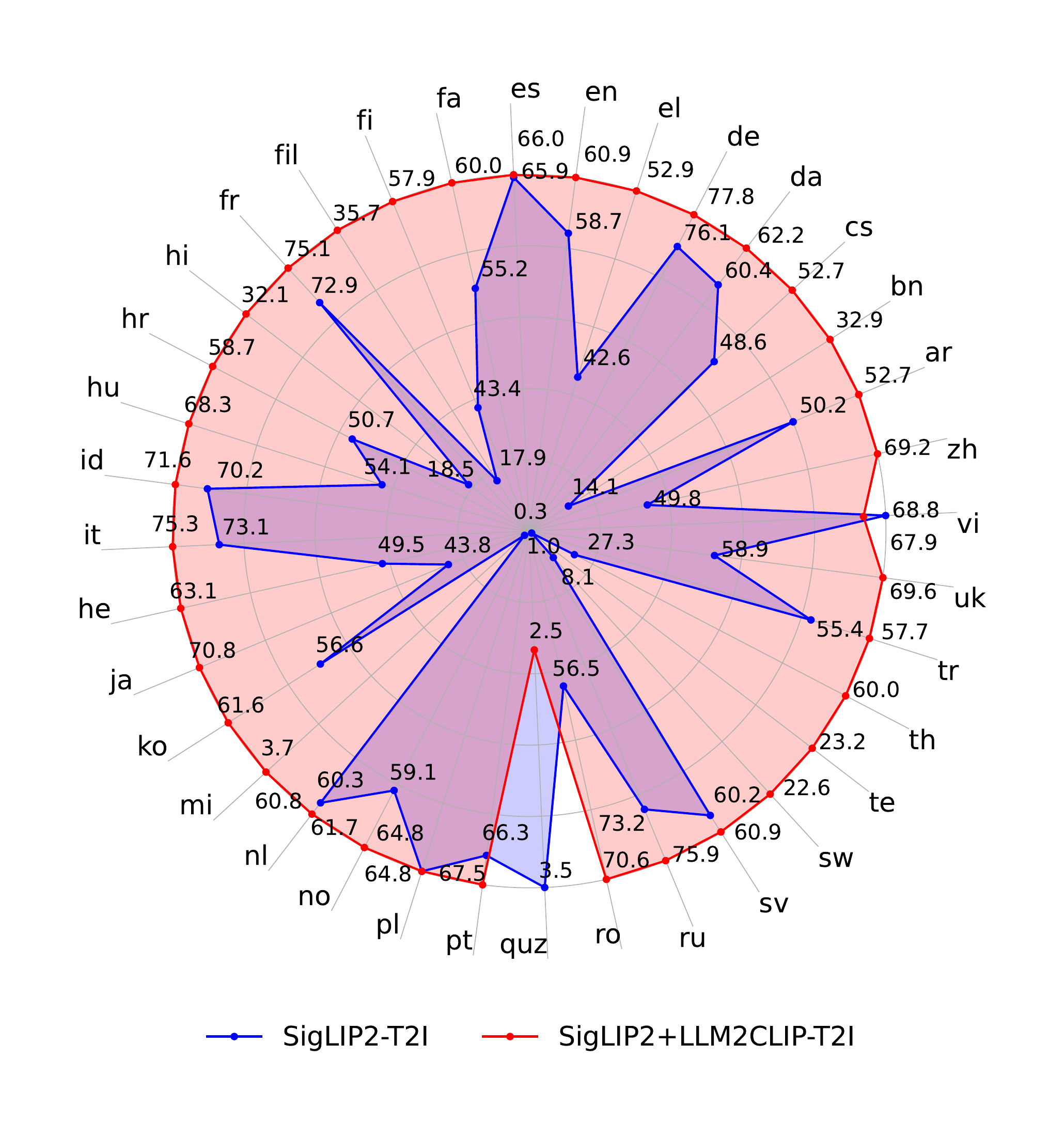}
        \caption{Multilingual zero-shot text-to-image retrieval (Recall@1)}
        \label{fig:xm_t2i}
    \end{subfigure}
    \caption{Multilingual zero-shot retrieval results on Crossmodal-3600.}
    \label{fig:xm_combined}
\end{figure*}

\begin{table*}[ht]
\centering
\renewcommand{\arraystretch}{1.2} 
\resizebox{0.8\textwidth}{!}{%
\begin{tabular}{l|cc|cc|cc|cc|cc|cc}
\shline
\multirow{2}{*}{\textbf{Methods(MLLM's Caption Ratio)}} & \multicolumn{2}{c|}{\textbf{Flickr}} & \multicolumn{2}{c|}{\textbf{COCO}} & \multicolumn{2}{c|}{\textbf{ShareGPT4V}} & \multicolumn{2}{c|}{\textbf{Urban-1k}} & \multicolumn{2}{c|}{\textbf{DOCCI}} & \multicolumn{2}{c}{\textbf{Average}} \\
                                                          & \textbf{I2T} & \textbf{T2I} & \textbf{I2T} & \textbf{T2I} & \textbf{I2T} & \textbf{T2I} & \textbf{I2T} & \textbf{T2I} & \textbf{I2T} & \textbf{T2I} & \textbf{I2T} & \textbf{T2I} \\
\shline
CLIP                         & 89.6 & 77.8 & 59.4 & 48.6 & 88.1 & 87.7 & 68.0 & 74.8 & 67.0 & 71.3 & 74.4 & 72.0 \\
Directly Finetune(50\%)       & 89.3 & 77.8 & 59.3 & 48.5 & 88.3 & 88.2 & 68.5 & 76.0 & 67.2 & 71.2 & 74.5 & 72.3 \\
\hline
\hspace{5mm}+~\ourmethod{} (100\%)           & 89.2 & 77.2 & 55.6 & 47.6 & 97.6 & \textbf{98.0} & \textbf{88.5} & \textbf{92.5} & 82.6 & 86.9 & 82.7 & 80.4 \\
\hspace{5mm}+~\ourmethod{} (75\%)            & \textbf{92.0} & 81.3 & 61.5 & 52.0 & \textbf{97.7} & 97.8 & 85.1 & 92.2 & \textbf{84.0} & \textbf{87.7} & \textbf{84.1} & \textbf{82.2} \\
\rowcolor[rgb]{0.925,0.957,1} \hspace{5mm}+~\ourmethod{} (50\%)            & 91.9 & 81.7 & \textbf{62.0} & 52.7 & 97.2 & 97.6 & 84.6 & 91.3 & 83.7 & 87.3 & 83.9 & 82.1 \\
\hspace{5mm}+~\ourmethod{} (25\%)            & 91.1 & \textbf{81.8} & 61.8 & \textbf{53.0} & 97.3 & 97.2 & 83.4 & 90.6 & 83.9 & 86.9 & 83.5 & 81.9 \\
\hspace{5mm}+~\ourmethod{} (0\%)             & 91.1 & 81.5 & 60.9 & \textbf{53.0} & 96.0 & 94.4 & 72.0 & 85.6 & 82.6 & 85.6 & 80.5 & 80.0 \\
\shline
\end{tabular}%
}
\caption{Ablation study on the proportion of dense captions from different MLLMs. Here, we explore the impact of both original short image captions and synthetic dense captions generated by MLLMs on the performance of~\ourmethod{}.}
\label{tab:stage-2-data}
\end{table*}

\begin{table*}[h]
\centering
\scriptsize 
\setlength{\tabcolsep}{1mm}
\begin{tabular}{l l c c | cc | cc | cc | cc | cc | cc}
\shline
\multirow{2}{*}{Model} & \multirow{2}{*}{Strategy} & \multirow{2}{*}{Training Hour} & \multirow{2}{*}{Batchsize} & \multicolumn{2}{c|}{Flickr} & \multicolumn{2}{c|}{COCO} & \multicolumn{2}{c|}{ShareGPT4V} & \multicolumn{2}{c|}{Urban1k} & \multicolumn{2}{c|}{DOCCI} & \multicolumn{2}{c}{Average} \\
 &  &  &  & I2T & T2I & I2T & T2I & I2T & T2I & I2T & T2I & I2T & T2I & i2t & t2i \\
\shline
Llama-3.1-8B      & LLM LoRA                                      & 17  & 768   & 90.0 & 80.9 & 60.9 & 50.6 & 97.5 & 96.9 & 88.9 & 90.8 & 82.0 & 84.3 & 83.9 & 80.7 \\
                  & LLM Frozen + Linear Adaptor + Offline-loading & 1.3 & 16384 & 85.2 & 74.6 & 52.3 & 40.8 & 87.7 & 89.6 & 57.5 & 59.2 & 49.9 & 48.3 & 66.5 & 62.5 \\
\shline
Llama-3.1-8B-CC  & LLM LoRA                                      & 17  & 704   & 90.5 & 81.8 & 61.5 & 52.4 & 98.1 & 98.2 & \textbf{91.0} & 92.8 & 85.7 & 87.4 & 85.4 & 82.5 \\
                  & LLM Frozen + Linear Adaptor                   & 5.5 & 4096  & 91.9 & 81.7 & 62.0 & 52.7 & 97.2 & 97.6 & 84.6 & 91.3 & 83.7 & 87.3 & 83.9 & 82.1 \\
                  & LLM Frozen + Linear Adaptor + Offline-loading & 1.3 & 16384 & \textbf{93.2} & \textbf{82.2} & \textbf{64.2} & \textbf{54.0} & \textbf{98.2} & \textbf{98.3} & 87.8 & \textbf{93.1} & \textbf{86.2} & \textbf{88.8} & \textbf{85.9} & \textbf{83.3} \\
\shline
\end{tabular}%
\caption{Efficiency study of~\ourmethod{}. We use two nodes, each equipped with 8 A100 40GB GPUs, and conduct experiments by maximizing GPU memory utilization to achieve the largest possible batch size. We report training hours, batch size, and retrieval accuracy for each experiment. During offloading, we precompute text captions offline into embeddings using the LLM. This approach is a one-time preprocessing step that effectively reduces the experimental overhead.}
\label{tab:eff}
\end{table*}

\begin{table*}[!h]
\centering
\footnotesize 
\setlength{\tabcolsep}{1mm}
\begin{tabular}{@{}l|cc|cc|cc|cc|cc|cc@{}}
\hline
\multirow{2}{*}{\textbf{Stage1}} & \multicolumn{2}{c|}{\textbf{Flickr}} & \multicolumn{2}{c|}{\textbf{COCO}} & \multicolumn{2}{c|}{\textbf{ShareGPT4V}} & \multicolumn{2}{c|}{\textbf{Urban-1k}} & \multicolumn{2}{c|}{\textbf{DOCCI}} & \multicolumn{2}{c}{\textbf{Avg}} \\
 & \textbf{I2T} & \textbf{T2I} & \textbf{I2T} & \textbf{T2I} & \textbf{I2T} & \textbf{T2I} & \textbf{I2T} & \textbf{T2I} & \textbf{I2T} & \textbf{T2I} & \textbf{I2T} & \textbf{T2I} \\
\hline
\rowcolor[cmyk]{0.05,0.02,0,0} Lora, AvgPool, Bidirectional, Supervise Simcse & 88.9 & \textbf{78.8} & 59.8 & \textbf{49.1} & \textbf{96.3} & 95.6 & \textbf{80.1} & 85.1 & \textbf{77.0} & \textbf{80.8} & \textbf{80.4} & \textbf{77.9} \\
Lora, AvgPool, Bidirectional, \textcolor[cmyk]{0,1,1,0}{Un-supervise Simcse} & 77.2 & 66.0 & 48.8 & 38.2 & 80.2 & 86.2 & 47.7 & 53.5 & 41.9 & 44.7 & 59.2 & 57.7 \\
Lora, AvgPool, Bidirectional, \textcolor[cmyk]{0,1,1,0}{MNTP} & 82.0 & 71.7 & 51.2 & 43.1 & 92.0 & 91.9 & 67.0 & 70.3 & 58.4 & 58.2 & 70.1 & 67.0 \\
Lora, AvgPool, Bidirectional, \textcolor[cmyk]{0,1,1,0}{MNTP}, Supervise Simcse & 88.2 & 78.3 & 59.6 & 48.9 & 95.5 & 95.1 & 79.7 & \textbf{86.1} & 75.3 & 77.7 & 79.7 & 77.2 \\
Lora, AvgPool, \textcolor[cmyk]{0,1,1,0}{Casual}, Supervise Simcse & 89.4 & 78.7 & \textbf{59.9} & 49.0 & 95.4 & \textbf{95.7} & 79.9 & 84.9 & 75.3 & 79.1 & 80.0 & 77.5 \\
Lora, \textcolor[cmyk]{0,1,1,0}{EOS}, Bidirectional, Supervise Simcse & \textbf{89.6} & 78.6 & 59.4 & 48.5 & 95.3 & 95.4 & 79.8 & 83.8 & 76.0 & 80.0 & 80.0 & 77.3 \\
\textcolor[cmyk]{0,1,1,0}{Frozen}, \textcolor[cmyk]{0,1,1,0}{Linear Adaptor}, AvgPool, Bidirectional, Supervise Simcse & 85.5 & 75.5 & 54.1 & 45.3 & 92.0 & 93.5 & 72.3 & 75.5 & 66.4 & 66.9 & 74.1 & 71.3 \\
\hline
\end{tabular}
\caption{Ablation study on the training methods of LLM caption contrastive finetuning in Stage 1. Rows with a light blue background represent our default setting, and red text indicates content that differs from the default setting.}
\label{tab:stage1_methods}
\end{table*}

\begin{table*}[!h]
\centering
\footnotesize
\begin{tabular}{@{}l|cc|cc|cc|cc|cc|cc@{}}
\hline
\multirow{2}{*}{\textbf{Method}} & \multicolumn{2}{c|}{\textbf{Flickr}} & \multicolumn{2}{c|}{\textbf{COCO}} & \multicolumn{2}{c|}{\textbf{ShareGPT4V}} & \multicolumn{2}{c|}{\textbf{Urban-1k}} & \multicolumn{2}{c|}{\textbf{DOCCI}} & \multicolumn{2}{c}{\textbf{Avg}} \\
 & \textbf{I2T} & \textbf{T2I} & \textbf{I2T} & \textbf{T2I} & \textbf{I2T} & \textbf{T2I} & \textbf{I2T} & \textbf{T2I} & \textbf{I2T} & \textbf{T2I} & \textbf{I2T} & \textbf{T2I} \\
\hline
CLIP & 89.6 & 77.8 & 59.4 & 48.6 & 88.1 & 87.7 & 68.0 & 74.8 & 67.0 & 71.3 & 74.4 & 72.0 \\
Directly Finetune (50\%) & 89.3 & 77.8 & 59.3 & 48.5 & 88.3 & 88.2 & 68.5 & 76.0 & 67.2 & 71.2 & 74.5 & 72.3 \\
\hline
bge-en-icl & 89.1 & 78.5 & 58.8 & 49.7 & 95.0 & 95.7 & 77.9 & 87.7 & 73.5 & 79.4 & 78.9 & 78.2 \\
LLM2Vec-Llama-3-8B & 91.1 & 81.4 & 61.5 & 51.9 & 94.5 & \textbf{96.4} & 82.1 & 88.6 & 77.7 & 82.6 & 81.4 & 80.2 \\
NV-Embed-v2 & 90.4 & 80.0 & 60.5 & 51.6 & 94.5 & 95.7 & 83.3 & \textbf{90.0} & 78.3 & 82.1 & 81.4 & 79.9 \\
\hline
bge-m3-XLM-R             & 80.7 & 70.3 & 51.4 & 42.0 & 84.5 & 86.5 & 56.8 & 63.4 & 51.7 & 55.9 & 65.0 & 63.6 \\
jina-v3-XLM-R            & 84.4 & 73.7 & 56.3 & 45.6 & 90.0 & 90.9 & 70.8 & 74.1 & 66.7 & 70.6 & 73.6 & 71.0 \\
e5 (XLM-R)               & 86.4 & 75.3 & 56.4 & 46.2 & 88.4 & 88.5 & 71.1 & 77.3 & 67.5 & 71.2 & 74.0 & 71.7 \\
\hline
VLM2VEC                  & 91.6 & 79.8 & 61.3 & 51.5 & 93.9 & 91.0 & \textbf{90.9} & \textbf{92.0} & 80.5 & 86.0 & 83.6 & 80.1 \\

VLM2VEC (finetune)       & 90.2 & 79.3 & 60.0 & 50.1 & 89.8 & 91.4 & 76.9 & 85.8 & 74.1 & 78.7 & 78.2 & 77.1 \\
\hline
Qwen2.5-0.5B-CC & 86.2 & 74.9 & 56.0 & 45.1 & 92.6 & 93.2 & 73.7 & 79.0 & 69.5 & 73.0 & 75.6 & 73.0 \\
Llama-3.2-1B-CC & 88.9 & 78.8 & 59.8 & 49.1 & 96.3 & 95.6 & 80.1 & 85.1 & 77.0 & 80.8 & 80.4 & 77.9 \\
Llama-3-8B-CC & 90.4 & 80.7 & 62.7 & 51.9 & 96.5 & 96.2 & 84.2 & 89.5 & 83.3 & \textbf{86.4} & 83.4 & 80.9 \\
DeepSeek-R1-Distill-Llama-8B-CC & 91.7 & 80.9 & 62.2 & 51.9 & 96.7 & 96.3 & 84.3 & 88.3 & 82.5 & 85.2 & 83.5 & 80.5 \\
Llama-3.1-8B-CC & \textbf{92.2} & \textbf{81.5} & \textbf{63.5} & \textbf{52.3} & \textbf{97.1} & 96.2 & \textbf{86.5} & 89.3 & \textbf{84.7} & 85.9 & \textbf{84.8} & \textbf{81.0} \\
\rowcolor[cmyk]{0.1,0.2,0,0}\textcolor[cmyk]{0,1,1,0}{Llama3.1-8B} & 85.2 & 74.6 & 52.3 & 40.8 & 87.7 & 89.6 & 57.5 & 59.2 & 49.9 & 48.3 & 66.5 & 62.5 \\
\hline
\end{tabular}
\caption{Ablation for using different text encoder. BGE-EN-ICL, LLM2Vec-Llama-3-8B, and NV-Embed-v2 are community-released text encoders derived from LLMs. The suffix "-CC" indicates encoders that have undergone our caption contrastive fine-tuning method based on the respective original LLMs. }
\label{tab:llm_encoders}
\end{table*}

\begin{table*}[ht]
\centering
\footnotesize
\begin{tabular}{l c | cc | cc | cc | cc | cc | cc}
\shline
\multirow{2}{*}{\textbf{Stage 1}} & \multirow{2}{*}{\textbf{Stage 2}} & \multicolumn{2}{c|}{\textbf{Flickr}} & \multicolumn{2}{c|}{\textbf{COCO}} & \multicolumn{2}{c|}{\textbf{ShareGPT4V}} & \multicolumn{2}{c|}{\textbf{Urban-1k}} & \multicolumn{2}{c|}{\textbf{DOCCI}} & \multicolumn{2}{c}{\textbf{Average}} \\
 &  & \textbf{I2T} & \textbf{T2I} & \textbf{I2T} & \textbf{T2I} & \textbf{I2T} & \textbf{T2I} & \textbf{I2T} & \textbf{T2I} & \textbf{I2T} & \textbf{T2I} & \textbf{I2T} & \textbf{T2I} \\
\shline
--           & --              & 89.2 & 77.7 & 59.0 & 47.3 & 94.3 & 94.3 & 75.0 & 80.5 & 74.0 & 77.5 & 78.3 & 75.5 \\
--           & Linear($\times1$)     & 88.5 & 77.5 & 59.8 & 48.4 & 95.1 & 94.5 & 77.2 & 84.4 & 75.3 & 78.7 & 79.2 & 76.7 \\
--           & Linear($\times2$)     & 89.6 & 78.1 & 59.6 & 48.5 & 95.5 & 94.7 & 79.7 & 84.0 & 76.1 & 79.0 & 80.1 & 76.8 \\
\rowcolor[rgb]{0.925,0.957,1}--           & Linear($\times4$)     & 88.9 & \textbf{78.8} & 59.8 & \textbf{49.1} & \textbf{96.3} & \textbf{95.6} & 80.1 & 85.1 & 77.0 & \textbf{80.8} & 80.4 & \textbf{77.9} \\
Linear($\times4$)   & Linear($\times4$)     & \textbf{89.9} & \textbf{78.8} & 59.7 & 48.9 & 95.1 & \textbf{95.6} & 79.9 & 84.9 & \textbf{78.0} & 80.4 & \textbf{80.5} & 77.7 \\
--           & Transformer($\times1$)& 88.6 & 78.7 & \textbf{59.9} & 48.6 & 95.8 & 95.1 & 79.5 & 84.5 & 77.0 & 79.8 & 80.2 & 77.3 \\
Transformer($\times1$)& Transformer($\times1$)& 89.8 & 78.6 & 58.1 & 49.0 & 96.1 & 95.4 & \textbf{81.9} & \textbf{85.6} & 76.5 & 78.1 & \textbf{80.5} & 77.3 \\
\shline
\end{tabular}%
\caption{Ablation experiments for adaptor design on Stage 1 caption contrastive finetune and Stage 2 \ourmethod{} post training. In parentheses, ($\times\textit{l}$) denotes an adaptor with $\times\textit{l}$ layers. The 4-layer Linear Adaptor has 67.1M parameters, while the single layer Transformer Adaptor has 67.6M parameters. This experiment is conducted using llama 3.1 1B and OpenAI CLIP ViT-L 224, with CC 3M data utilized in both Stage 1 and Stage 2. The light blue row indicate our default setting.}
\label{tab:adaptor_analysis}
\end{table*}

\begin{table*}[!h]
\centering
\scriptsize              
\setlength{\tabcolsep}{1mm}
\begin{tabular}{l c | c c | c c | c c | c c | c c | c c | c c}
\hline
\multicolumn{2}{c|}{\multirow{2}{*}{\textbf{Methods}}} & \multirow{2}{*}{\textbf{Training Loss}} & \multirow{2}{*}{\textbf{Testing Text Encoder}} & \multicolumn{2}{c|}{\textbf{Flickr}} & \multicolumn{2}{c|}{\textbf{COCO}} & \multicolumn{2}{c|}{\textbf{ShareGPT4V}} & \multicolumn{2}{c|}{\textbf{Urban1k}} & \multicolumn{2}{c|}{\textbf{DOCCI}} & \multicolumn{2}{c}{\textbf{Average}} \\
 &  &  &  & \textbf{I2T} & \textbf{T2I} & \textbf{I2T} & \textbf{T2I} & \textbf{I2T} & \textbf{T2I} & \textbf{I2T} & \textbf{T2I} & \textbf{I2T} & \textbf{T2I} & \textbf{I2T} & \textbf{T2I} \\
\hline
CLIP              &               & CL(CLIP-T, CLIP-V) & CLIP-T & 89.6 & 77.8 & 59.4 & 48.6 & 88.1 & 87.7 & 68.0 & 74.8 & 67.0 & 71.3 & 74.4 & 72.0 \\
Directly Finetune &               & CL(CLIP-T, CLIP-V) & CLIP-T & 89.3 & 77.8 & 59.3 & 48.5 & 88.3 & 88.2 & 68.5 & 76.0 & 67.2 & 71.2 & 74.5 & 72.3 \\
\hline
\multirow{8}{*}{\hspace{3mm}+\ourmethod{}} 
                  & {a)} & CL(LLM, CLIP)      & LLM    & \textbf{91.9} & 81.7 & 62.0 & 52.7 & 97.2 & 97.6 & 84.6 & 91.3 & 83.7 & 87.3 & 83.9 & 82.1 \\
\cline{3-16}
                  & \multirow{2}{*}{b)} & \multirow{2}{*}{\begin{tabular}[c]{@{}c@{}}CL(LLM, CLIP-V)+\\CL(CLIP-T, CLIP-V)\end{tabular}} & CLIP-T & 89.6 & 77.8 & 59.4 & 48.6 & 88.1 & 87.7 & 68.0 & 74.8 & 67.0 & 71.3 & 74.4 & 72.0 \\
                  &                   &                                          & LLM    & 90.1 & 81.3 & 61.3 & 52.4 & 97.1 & 97.4 & 85.6 & 91.1 & 83.8 & 86.9 & 83.6 & 81.8 \\
\cline{3-16}
                  & \multirow{2}{*}{c)} & \multirow{2}{*}{\begin{tabular}[c]{@{}c@{}}CL(LLM, CLIP-V)+CL(CLIP-T, CLIP-V)+\\CL(CLIP-T, LLM)\end{tabular}} & CLIP-T & 89.0 & 77.9 & 58.5 & 47.7 & 89.1 & 88.8 & 67.0 & 75.5 & 66.3 & 70.7 & 74.0 & 72.1 \\
                  &                   &                                          & LLM    & 90.8 & 80.5 & 61.5 & 51.8 & 97.1 & 97.7 & 85.9 & 90.3 & 83.4 & 86.7 & 83.7 & 81.4 \\
\cline{3-16}
                  & \multirow{3}{*}{d)} & \multirow{3}{*}{\begin{tabular}[c]{@{}c@{}}CL(LLM, CLIP-V)+CL(CLIP-T, CLIP-V)+\\CL(Cat(CLIP-T, LLM), CLIP-V)\end{tabular}} & CLIP-T & 89.6 & 77.9 & 59.1 & 48.1 & 89.4 & 87.5 & 69.1 & 72.5 & 67.0 & 70.6 & 74.8 & 71.3 \\
                  &                   &                                          & LLM    & 90.7 & 81.4 & 61.3 & 52.3 & 97.5 & \textbf{97.9} & 85.5 & 92.5 & 83.9 & 87.5 & 83.8 & 82.3 \\
                  &                   &                                          & Cat(CLIP-T,LLM) & 91.3 & \textbf{82.4} & \textbf{62.9} & \textbf{53.3} & \textbf{97.8} & 97.4 & \textbf{86.6} & \textbf{92.9} & \textbf{84.9} & \textbf{88.2} & \textbf{84.7} & \textbf{82.8} \\
\cline{3-16}
\hline
\end{tabular}
\caption{\ourmethod{} Training Method Analysis. We experimented with various possibilities for fine-tuning between the LLM and the pretrained CLIP model's Vision Encoder (CLIP-V) and Text Encoder (CLIP-T). In the figure, "CL" denotes the contrastive learning loss, and "Cat" stands for concatenation. Items a--d correspond to the training methods section in~\Cref{sec:method_stage2}. In this experiment, we performed zero-shot image-text retrieval tests on each text encoder, including the new feature projection space formed after concatenation.}
\label{tab:stage-2-methods} 
\end{table*}

\paragraph{Detailed Results for X3600 Multilingual Task.}
In Figure~\ref{fig:xm_i2t} and Figure~\ref{fig:xm_t2i}, we visualize the I2T and T2I results for 36 different languages. It can be observed that our method outperforms the original SigLip2 text encoder, despite the latter being trained with 10\% of its training samples from 12 billion alt-texts covering 109 languages. In contrast, our method~\ourmethod{} was trained exclusively on 15 million English-only samples. This clearly demonstrates the inherent multilingual advantage gained by incorporating an LLM.

\paragraph{Data Ratio Ablation.}
In ~\Cref{tab:stage-2-data}, we experiment with various dataset compositions. The results show a clear trend: higher proportions of dense captions enhance performance in long-text image retrieval tasks, whereas lower dense caption ratios improve short-text retrieval performance. We select a 1:1 ratio by default, as it provides the most balanced improvement for both short and long texts.

\textit{Interestingly, even training exclusively on real short captions (100\%), our model still significantly boosts performance for both short and long-text retrieval tasks.} This improvement can be attributed to the incorporation of LLM, given that real captions share the same distribution as the CLIP pre-training data, yet LLM integration still yields substantial gains.

However, when training exclusively on synthesized long texts, although retrieval performance on long-text tasks can be significantly improved, performance on short-text retrieval tasks tends to decline. We hypothesize two possible reasons for this phenomenon: First, relying entirely on synthesized data might introduce significant biases in the training samples. Second, when the learned feature space is constructed exclusively from long texts, the model may prioritize global semantic matching and overlook fine-grained information such as specific words, potentially negatively affecting short-text retrieval. Therefore, we believe that for contrastive learning tasks focusing solely on long texts, it is important to consider including both global and local levels of text granularity during training like FLIP~\cite{yao2021filip}, or introducing sufficient hard examples to encourage the model to attend to low-level semantic details.

\paragraph{Optimization Strategy Ablation and Efficiency Analysis of \ourmethod{}.}
As shown in ~\Cref{tab:eff}, we conducted experiments comparing methods such as LoRA training, frozen LLM online training, and offloading on fixed hardware. Experimental results demonstrate that, with identical computational resources, our proposed offloading method can increase the batch size by tenfold and extend training time by fourfold, ultimately achieving higher performance. 

\paragraph{Complete Tables and Additional Visualizations.}
In Tables~\Cref{tab:stage1_methods},~\Cref{tab:llm_encoders},~\Cref{tab:adaptor_analysis}, and~\Cref{tab:stage-2-methods}, we provide additional detailed results corresponding to Table 6, 7, 8, and 9 in the main text, facilitating a clear analysis of the differences in retrieval performance between long and short texts. The conclusions remain consistent with those presented in the main text. Additionally,~\Cref{fig:radar} illustrates radar plots highlighting the performance improvements achieved by~\ourmethod{} on SigLip2 and EVA02.

\end{document}